\documentclass{article}
\usepackage{iclr2027_conference,times}
\usepackage{hyperref}
\usepackage{url}
\usepackage{booktabs}
\usepackage{array}
\usepackage{amsmath,amssymb}
\usepackage{graphicx}
\usepackage{placeins}
\usepackage{xcolor}
\usepackage{microtype}

\title{Shared Weights, Selected Computations:\\How Looped Transformers Route\\What Each Loop Does}
\author{Jiaju Wu$^{1,2}$ \qquad Yi Hu$^1$ \qquad Muhan Zhang$^1$}
\iclrfinalcopy
\hypersetup{pdfauthor={Jiaju Wu, Yi Hu, Muhan Zhang},pdftitle={Shared Weights, Selected Computations: How Looped Transformers Route What Each Loop Does}}

\usepackage{subcaption}
\begin{document}
\maketitle
\begingroup
\makeatletter
\let\@footnotetext\H@@footnotetext
\makeatother
\renewcommand{\thefootnote}{}
\footnotetext{\textsuperscript{1}Institute for Artificial Intelligence, Peking University. \textsuperscript{2}School of Mathematical Sciences, Peking University. Correspondence to: Muhan Zhang \textless\href{mailto:muhan@pku.edu.cn}{muhan@pku.edu.cn}\textgreater.}
\endgroup
\fancyhead[L]{Preprint}

\begin{abstract}
Looped Transformers repeatedly apply the same set of Transformer layers, giving them a recurrent architecture for latent computation. Their strong performance on iterative reasoning and length-generalization tasks suggests an appealing explanation: recurrence may provide an inductive bias that lets the model reuse a learned algorithm across loops. However, weight sharing alone does not imply that every loop performs the same operation. This raises a basic question: \textbf{is each loop actually repeating the same computation, and if not, what routes the shared parameters to different operations?}
We study this question using graph walks as a test case. In the model's native trajectories, decoded predictions can advance by different numbers of graph steps or remain at a reached target, showing that recurrent progress need not follow a fixed one-loop-one-step pattern. We then show that a frozen loop can be steered toward different transitions by modifying its entering hidden state: a learned linear layer $J$ selects the desired transition without changing the shared Transformer layers.
To test how this steering works, we use activation patching and find that attention patterns can recover its effects and switch the selected transition. Across five matched pairs of graph models, changing intermediate supervision during backbone training changes which transitions $J$ can induce. This suggests that $J$ selects computations learned by the backbone rather than creating new algorithms. Together, these results show that the hidden state can control shared computation, with attention routing as a causal pathway.
\end{abstract}

\begin{center}
\small Code and reproduction materials: \url{https://github.com/wjjpku/howloop}
\end{center}

\section{Introduction}
\label{sec:intro}

Recent work on \emph{latent reasoning} studies how models can carry out multi-step reasoning within their hidden states, without generating every intermediate step as text \citep{saunshi2025looped,geiping2025scaling}.
Recurrent architectures provide a natural way to support this kind of computation: the same parameters can be applied repeatedly while the internal state continues to evolve.
Looped Transformers bring this idea to Transformer models \citep{vaswani2017attention} by reusing the same set of Transformer layers over multiple loops \citep{dehghani2019universal,yang2024learning}.
This recurrent design provides a useful inductive bias for iterative computation.
Looped Transformers can learn iterative learning algorithms and multi-step optimization procedures \citep{yang2024learning,gatmiry2024gradient}, improve length generalization on tasks with iterative solutions \citep{fan2025looped}, and use additional loops as a source of latent test-time computation at language-model scale \citep{geiping2025scaling,zhu2025ouro}.
These results suggest that recurrent depth can provide a way to reuse learned computation over multiple internal steps.

\begin{figure}[!b]
  \centering
  \includegraphics[pagebox=cropbox,width=\linewidth]{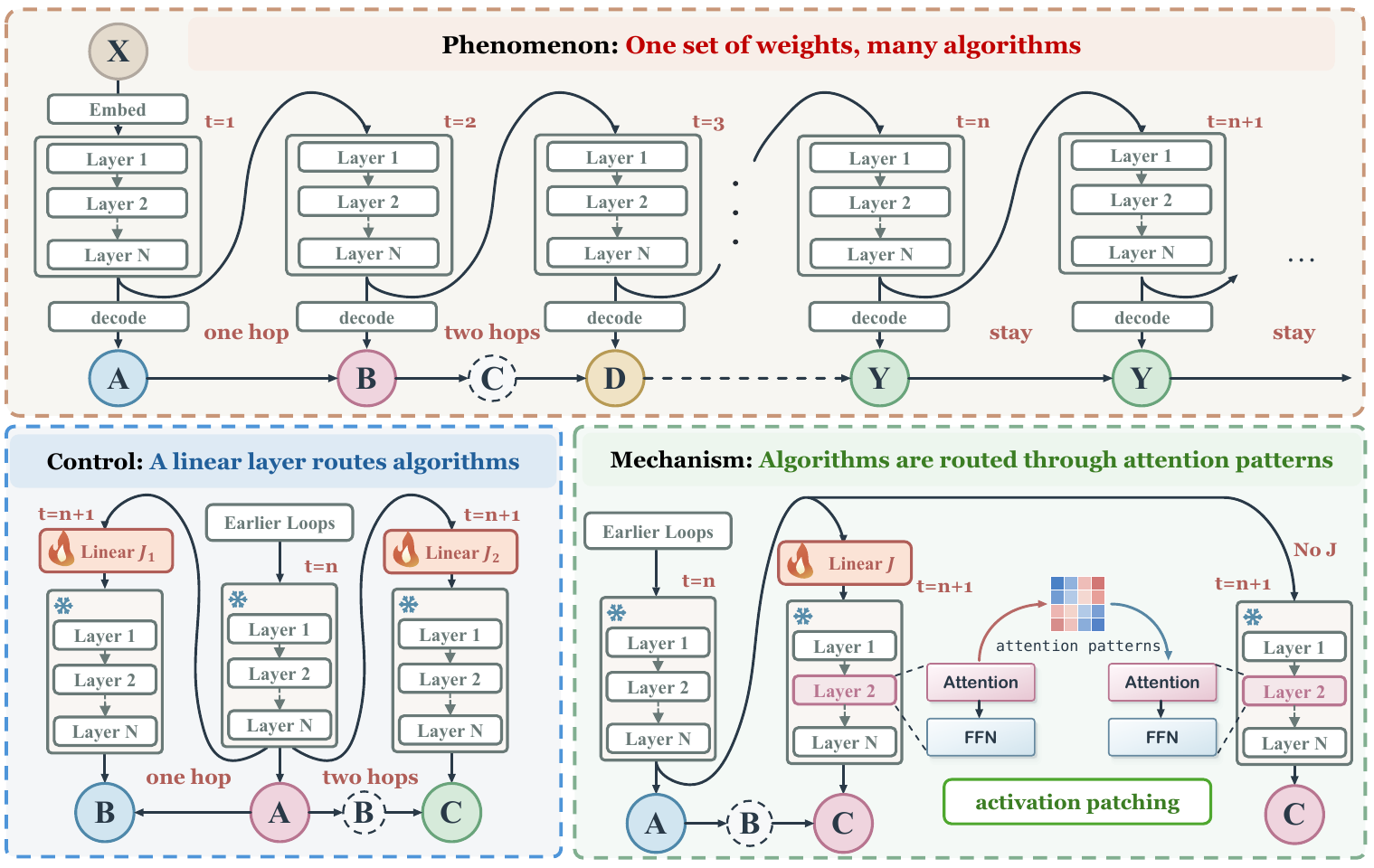}
  \caption{Overview of looped computation and state steering. \textbf{Phenomenon:} intermediate readouts reveal different patterns of progress across weight-shared loops. \textbf{Control:} a learned linear layer changes the state entering the next loop to select a different transition while the backbone remains frozen. \textbf{Mechanism:} patching attention patterns from a steered run into an unsteered run transfers routing while preserving the receiving run's values, recovering steering effects. Flames mark trainable components; snowflakes mark frozen components.}
\label{fig:mechanism}
\end{figure}

These results naturally suggest an iterative view of Looped Transformers: the shared layers may repeatedly apply the same learned procedure as computation progresses.
For example, in a graph walk, one might expect every loop to move one step along the path.
Yet weight sharing does not imply such a fixed correspondence.
The parameters are shared across loops, but the hidden state entering those parameters changes as computation proceeds.
The same network is therefore applied to different internal states, so weight sharing alone does not specify what operation each loop performs.
The training objective can also leave the intermediate trajectory underdetermined.
When only the final answer is supervised, the loss does not require the model to follow any particular sequence of intermediate steps.
This leads to a basic question:
\emph{is each loop actually repeating the same computation, and if not, what routes the shared parameters to different operations?}
Answering this question is important for understanding the inductive bias of recurrent depth and the internal process of latent reasoning.
It also affects when additional loops provide useful computation and whether recurrent test-time computation can be controlled.

Prior work has gradually moved from asking whether recurrence can implement algorithms to studying how learned computation is organized across loops.
Constructive results show that fixed-weight Looped Transformers can execute multi-step programs, while learning-based analyses show that looped models can learn iterative optimization procedures \citep{giannou2023loops,yang2024learning,gatmiry2024gradient}.
These results establish that shared parameters can support reusable algorithmic computation, but they do not specify how that computation is distributed across loops after training.
Mechanistic work therefore turns to the learned recurrent trajectory itself.
\citet{blayney2026mechanistic} identify layer-wise fixed points, cyclic latent trajectories, and repeated stages of inference across loops.
Probing studies likewise find that intermediate recurrent states do not always form a simple, directly readable latent chain of thought \citep{lu2025probing}.
More recently, \citet{zhang2026recurrence} study task progress across loops directly.
They show that training can select computation frontiers with different speeds, so the amount of progress associated with one loop is learned rather than fixed by weight sharing.
Together, these studies characterize how recurrent computation is learned and organized across loops, but leave open whether the state entering a loop can causally select what the same frozen layers compute next, and how this selection is implemented inside the model.

We address these gaps by testing whether the entering hidden state can redirect the next computation of a frozen loop, and by tracing this control to attention routing (Figure~\ref{fig:mechanism}).
We begin with the recurrent trajectories that arise without intervention.
In graph-walk tasks, native readouts can advance at different numbers of graph steps or remain at an already reached target.
A learned trajectory therefore need not follow a fixed ``one loop = one algorithmic step'' rule (Section~\ref{sec:underdetermined}).
This variation makes the entering hidden state a natural candidate for controlling the next computation.
We freeze the shared Transformer layers and learn a linear layer at the loop boundary.
Changing only the entering state selects one-hop or two-hop targets for the same frozen loop (Section~\ref{sec:target}).
Having established this control, we ask how it is implemented inside the loop.
Patching attention patterns recovers much of the steering effect in the graph case study and the Ouro multi-hop language task. Patching patterns between one-hop and two-hop runs also switches the semantic target, providing causal evidence that routing carries the choice of transition (Section~\ref{sec:hypothesis}).
Finally, we test how backbone training and repeated execution shape this control.
Across five matched pairs of graph models, intermediate supervision during backbone training changes which computations $J$ can steer. This supports the view that $J$ controls computations learned by the backbone rather than creating new algorithms (Section~\ref{sec:generalization}). The controllers can be composed across two successive loops, but accuracy is order-dependent and drops with longer compositions (Section~\ref{sec:composition}).

\section{Related Work}
\label{sec:related}

\paragraph{Recurrent computation.}
Recurrent architectures increase computation depth by repeatedly applying the same parameters \citep{dehghani2019universal}. Prior work shows that looped Transformers can execute programs and learn iterative algorithms \citep{giannou2023loops,yang2024learning,gatmiry2024gradient}, improve length generalization \citep{fan2025looped}, and use additional loops to improve reasoning \citep{saunshi2025looped,geiping2025scaling,zhu2025ouro}. These results establish the value of recurrent computation. They do not, however, determine what operation each loop performs in a trained model.

\paragraph{Computation learned across loops.}
Understanding loop behavior requires distinguishing the algorithms an architecture can express from the computation selected by training. Training budgets can select different rates of task progress within the same shared architecture \citep{zhang2026recurrence}. Studies of looped language models also identify cyclic latent trajectories and repeated stages of inference \citep{blayney2026mechanistic}. Intermediate readouts need not directly reveal this internal computation: their interpretation can depend on the layer and decoding method \citep{lu2025probing}. These findings motivate going beyond observing an existing trajectory. We ask whether changing the hidden state entering a loop can make the same parameters perform different operations on the input.

\paragraph{Supervision of recurrent states.}
Supervision provides a way to examine how learned computation affects state control. Training objectives shape both the recurrent trajectory and the information available at intermediate readouts \citep{fan2026lotus,popescu2026adaptive}. Supervising outputs does not fully specify the internal computation; even per-loop supervision can leave state variables such as hidden-state scale uncontrolled \citep{sharma2026readout}. We therefore compare backbones trained under different supervision schemes to test whether steering depends on computational capabilities already learned by the backbone.

\paragraph{State steering and causal interventions.}
State interventions and activation patching provide complementary tools for testing this account. Steering modifies hidden states to control a frozen model's behavior \citep{turner2023activation}, including through learned affine maps \citep{singh2024surgery}. Activation patching replaces internal activations to test which components contribute to a behavior \citep{wang2023ioi}. We combine these tools by using an affine map to change the loop input and select the next target, then patching attention patterns to test whether routing carries this selection. This extends the study of what recurrence can compute to whether learned computation can be selected through the entering state and how that selection is implemented.

\section{Phenomenon: Task Progress Varies across Loops}
\label{sec:underdetermined}
An iterative view of looped Transformers suggests that shared layers may repeatedly apply the same learned algorithm across loops. We examine this view using readouts at intermediate loops on the graph walk task to observe how computation progresses as the model runs.

Each input consists of a permutation $f_G$ on a graph $G$, a start node $s$, and a requested path depth $d = 8$. Writing $f_G(v)$ for the successor of $v$, the target is $f_G^8(s)$. We train two models, D8L8 and D8L6, which repeat a two-layer Transformer block for eight and six loops, respectively. Only the final answer is supervised.

\begin{figure}[!t]
\centering
\includegraphics[width=\linewidth]{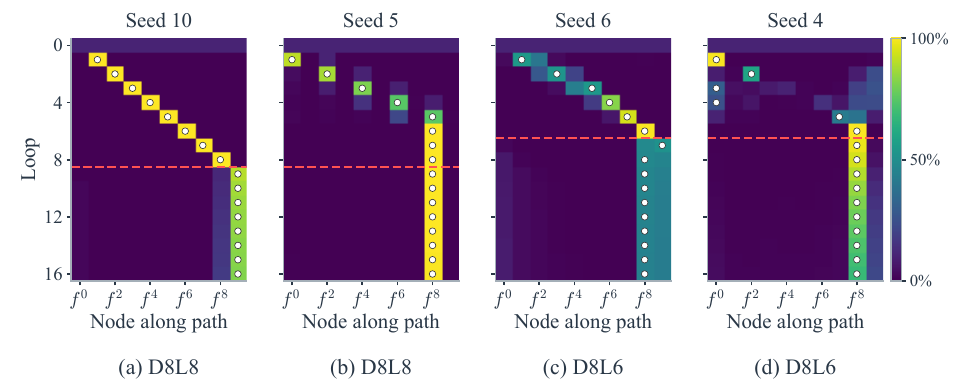}
\caption{Color shows prediction frequency. Red dashed lines mark the training loop $L = 8$ or $L = 6$; white dots mark the unique most frequent prediction at each loop.}
\label{fig:trajectory}
\end{figure}

For visualization, we take the trained looped block, repeat it, and apply the final readout head after loops 0--16. We intentionally use ten-node cycles for testing so that the iteration length can be clearly identified from the node index. Each plotted block in Figure~\ref{fig:trajectory} summarizes predictions across examples for each individual walk. We show selected representative trajectories; Appendix~\ref{app:all-native-trajectories} provides all twelve D8L8 and five D8L6 trajectories.

As shown in Figure~\ref{fig:trajectory}, the models can reach the same target through different readout trajectories. For D8L8, one trajectory advances mainly one node per loop, whereas another advances mainly two nodes per loop. For D8L6, the model must fit the advancement process into six loops, thus producing trajectories with irregular speeds. But eventually they stabilize at the final answer, even if additional loops are run. This shows that, within a single model, the algorithm can differ across loops. Some loops advance quickly, others slowly, and all eventually stabilize at the end. Sharing weights does not guarantee the same algorithm.

Before selecting models for mechanistic analysis, we must inspect these readouts. These trajectories reveal which mechanism the model primarily uses. We next turn this observed variation into a control test.

\section{Control: State Steering Redirects the Next Target of a Frozen Loop}
\label{sec:target}
The different algorithms observed above motivate us to study why they differ. To this end, we try to control the next transition by modifying the hidden state entering the block. We train a linear layer to control the next loop's target while keeping the backbone weights frozen. We use graph walks as a test case for steering the computation through the state entering the loop. Additional state-control experiments and supporting evidence on Qwen and synthetic knowledge-graph relation composition are provided in Appendix~\ref{app:additional-control}.

\subsection{State Steering with a Frozen Backbone}
Let $F$ denote the shared Transformer block and $h_t$ the sequence of token states after loop $t$. Our original model computes $h_{t+1}=F(h_t)$. To change the next transition, we apply a linear layer $J$ before the next loop, which acts independently on every token:
\begin{equation}
 h_{t+1}=F(J(h_t)).
\end{equation}
We train $J$ while keeping the backbone frozen. In the graph experiments, $J$ consists of a learned diagonal term, a rank-48 term, and a bias. The token-wise transformation, linear form, and low-rank parameterization constrain the expressive power of $J$. In particular, $J$ cannot aggregate graph information across token positions on its own; the frozen block $F$ must execute that part of the transition.

We evaluate the fitted map on out-of-distribution graphs that were unseen during both backbone training and map training. This tests whether the map generalizes to unseen graphs and whether modifying the information already represented in each token can actually redirect computation. Appendix~\ref{app:main-config} details the graph setup, controller training, and composition evaluations.

We first test whether two maps can select different next targets from the same state. D8L6 seed 6 processes an eight-hop query for six loops to produce $h_6$. Let $u=f_G^8(s)$ be the correct answer to this query. We train $J_{\mathrm{one}}$ so that one additional loop $F(J_{\mathrm{one}}(h_6))$ predicts $f_G(u)$. We train $J_{\mathrm{two}}$ in the same way, with target $f_G^2(u)$.

Both maps select their intended targets on held-out graphs (Figure~\ref{fig:target-behavior}). Without steering, one-hop accuracy is 48.2\%, but with $J_{\mathrm{one}}$ it rises to 99.6\%. Two-hop accuracy reaches 98.2\% with $J_{\mathrm{two}}$. Thus, changing the entering state selects different next targets under the same frozen weights and one additional loop.

\begin{center}\begin{minipage}{\linewidth}
\captionsetup{type=figure}
\centering
\includegraphics[width=.80\linewidth]{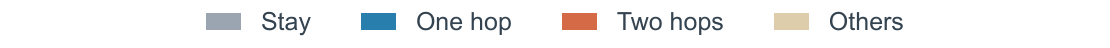}\par\smallskip
\captionsetup[subfigure]{font=normalsize,hypcap=true,labelformat=parens,labelsep=space,justification=centering}
\setcounter{subfigure}{0}%
\begin{subfigure}[t]{.32\linewidth}
\includegraphics[width=\linewidth]{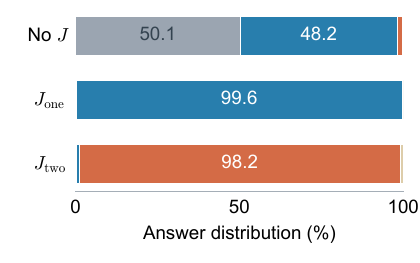}
\caption{from $h_6$}\label{fig:target-behavior}
\end{subfigure}\hfill
\begin{subfigure}[t]{.32\linewidth}
\includegraphics[width=\linewidth]{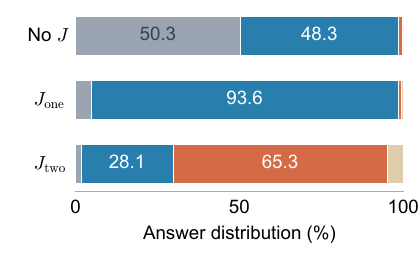}
\caption{from $F(J_{\mathrm{one}}(h_6))$}\label{fig:composition-one}\label{fig:composition}
\end{subfigure}\hfill
\begin{subfigure}[t]{.32\linewidth}
\includegraphics[width=\linewidth]{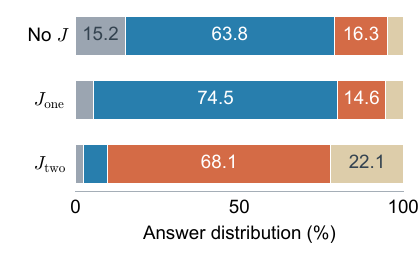}
\caption{from $F(J_{\mathrm{two}}(h_6))$}\label{fig:composition-two}
\end{subfigure}
\caption{One-loop answer distributions for D8L6. Each row corresponds to a map applied to the labeled starting state before $F$. In panels (a), (b), and (c), colors are referenced to $u$, $f_G(u)$, and $f_G^2(u)$, respectively. Bars show the average over two map fits.}
\label{fig:target}
\end{minipage}\end{center}

\begin{figure}[!b]
\centering
\includegraphics[width=.72\linewidth]{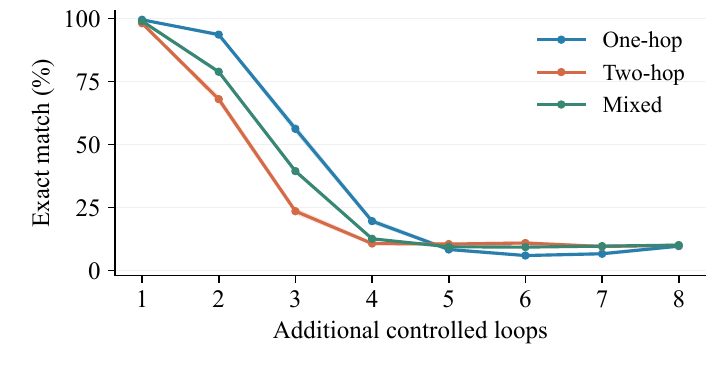}
\caption{Controller reuse on 512 new held-out graphs with D8L6 seed 6. The horizontal axis counts additional controlled loops after $h_6$; exact match compares the predicted node with the cumulative target at each call. Curves average two map fits on the same 3,175 examples; mixed averages 32 fixed sequences. Shading shows 95\% graph-bootstrap intervals.}
\label{fig:long-composition}
\end{figure}

\subsection{Composing Controller J}
\label{sec:composition}
Although each map is trained for a single additional loop, it can continue to control the next target when reused on a state produced by a previous controlled loop. We first evaluate all four two-map orders, then test repeated control for up to eight additional calls.

Applying the one-hop map twice reaches the two-hop target in 93.6\% of examples. Omitting the second map reduces accuracy to 48.3\%. Applying the one-hop map followed by the two-hop map reaches the three-hop target with 65.3\% accuracy. Reversing their order reaches the same target with 74.5\% accuracy (Figure~\ref{fig:target}). Thus, the maps can be reused over two loops, but their order affects accuracy.

We separately test longer reuse on 512 new held-out graphs, retaining 3,175 examples with distinct targets through four graph hops. Figure~\ref{fig:long-composition} follows repeated one-hop, repeated two-hop, and mixed maps for eight additional controlled loops after $h_6$. The target advances by the cumulative number of requested hops. We score exact match between the predicted node and this cumulative target at each call. At the third call, repeated one-hop and two-hop maps reach 56.3\% and 23.6\% accuracy. At the eighth, mixed exact match is 10.1\%. Thus, control persists beyond the single-loop training setting but degrades under longer reuse. We next intervene inside the frozen block to study how this control works.

\section{Mechanism: Attention Routing Mediates State-Dependent Computation}
\label{sec:hypothesis}\label{sec:large}
State steering changes the next target, but it does not reveal how the frozen block uses the changed state. We first identify how attention patterns route retrieved content. We then test whether attention-pattern patching produces the same steering effect as the linear map $J$. Finally, we directly switch the selected transition by exchanging patterns between one-hop and two-hop controlled runs.

To make steering cleaner, we need backbones that use both one-hop and two-hop algorithms. We therefore examine the readout at intermediate loops and select backbones whose trajectories advance at different speeds. For each backbone, we average two map fits, and then average the three backbone results with equal weights.

To test whether the mechanism extends to a large language model, we also evaluate Ouro-2.6B on a multi-hop language task. The model reads sentences describing successive letter transfers and answers who holds the letter after a requested number of transfers. We lightly fine-tune Ouro on 1--4-step requests with four loops. Then we freeze the backbone and train a dense linear layer on 1--8-step requests. The layer is applied before loops 2--4.

The three D8L6 interventions use different sets of activations. The pattern-versus-output experiment patches one selected head in the second layer at the answer position (Figure~\ref{fig:parallel-mechanism}b). The steering-transfer experiment patches the attention patterns of all four second-layer heads at that same position (Figure~\ref{fig:parallel-mechanism}c). The target-switching experiment patches the patterns of all four second-layer heads at every token position (Figure~\ref{fig:target-pattern-patching}). Ouro uses a fixed set of sixteen selected heads at its fourth loop. Head selection and evaluation use separate data. Appendices~\ref{app:main-config} and~\ref{app:parallel} provide the full setups, intervention protocols, and results for D8L6 and Ouro, respectively.

\subsection{Pattern and Output Patches Select Different Answers}
\label{sec:causal}
We patch activations from run 1 into run 2 to separate attention routing from retrieved content. The inputs have aligned token positions. Writing $\alpha_i$ for attention patterns and $V_i$ for values, the original head output in run 2 is $z_{\mathrm{raw}}=\alpha_2V_2$. We compare:
\begin{equation}
\begin{aligned}
z_{\mathrm{raw}} &\xrightarrow{\text{pattern patch}} z_{\mathrm{pattern}}=\alpha_1V_2,\\
z_{\mathrm{raw}} &\xrightarrow{\text{output patch}} z_{\mathrm{output}}=z_1=\alpha_1V_1.
\end{aligned}
\end{equation}
In D8L6, both runs use $J_{\mathrm{one}}$. Suppose run 1 starts the controlled transition at C with C$\to$D, while run 2 starts at A with A$\to$B and C$\to$E. Their original answers are D and B. Patching run 1's pattern into run 2 should select its C record but read E, changing B$\to$E. Patching the output should instead transfer D, changing B$\to$D.

For Ouro, the rule sets share seven transfers from $s_1$: $s_1\xrightarrow{7}\mathrm{C}\to\mathrm{D}$ in run 1 and $s_1\xrightarrow{7}\mathrm{C}\to\mathrm{E}$ in run 2. Run 2's query instead follows $s_2\xrightarrow{8}\mathrm{B}$. We again expect B$\to$E from pattern patching and B$\to$D from output patching. We require distinct B, D, E; D8L6 also requires correct pre-patch current-node and next-node readouts, while Ouro retains all 256 constructed pairs.

\begin{figure}[!t]
\centering
\includegraphics[width=\linewidth]{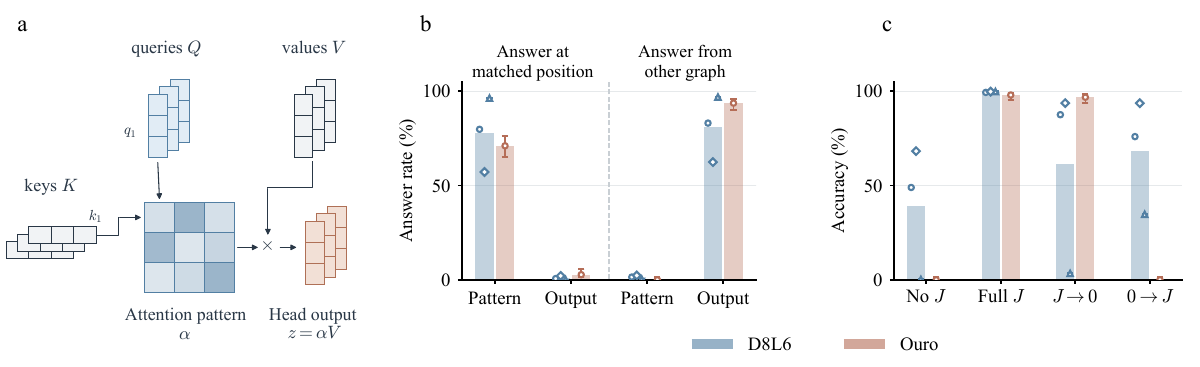}
\caption{Attention routing and state steering. (a) Attention computation. (b) Attention-pattern and output patches between graphs. (c) Attention-pattern patches between steered ($J$) and unsteered ($0$) runs. D8L6 bars are averaged over backbone means; markers indicate seeds 6 (circle), 10 (diamond), and 13 (triangle). Ouro error bars show 95\% Wilson intervals.}
\label{fig:parallel-mechanism}\label{fig:causal}
\end{figure}

The results follow these predictions (Figure~\ref{fig:parallel-mechanism}b). Pattern patches produce E in 77.8\% of D8L6 cases and 71.1\% of Ouro cases; output patches produce D in 80.9\% and 93.8\%, respectively. Thus, attention patterns can steer which content is selected while preserving the patched run's values.

\subsection{Attention Patterns Carry State-Steering Effects}
\label{sec:routing}\label{sec:qkv}
We next test whether changing the attention pattern alone can reproduce the effect of $J$. We patch attention patterns between steered and unsteered runs on the same input. D8L6 uses the one-hop map. Ouro is evaluated on eight-hop requests in the language task. Each run keeps its own values and steering schedule; only the attention pattern changes:
\begin{equation}
 z_{J\rightarrow0}=\alpha_J V_0,\qquad
 z_{0\rightarrow J}=\alpha_0 V_J.
\end{equation}
Here $J\rightarrow0$ patches the steered pattern into the unsteered run, and $0\rightarrow J$ does the reverse. If the steering effect is carried by the attention pattern, the first patch should improve unsteered accuracy, while the second should reduce steered accuracy.

The mean effects follow this prediction (Figure~\ref{fig:parallel-mechanism}c). In D8L6, patching steered patterns into the unsteered run raises one-hop accuracy from 39.2\% to 61.6\%. The reverse patch reduces accuracy from 99.7\% to 68.1\%.

The effect on the unsteered run varies across backbones, but on average the patch still improves accuracy. A single selected head can already transfer part of this effect: for seed 10, patching head 0 raises accuracy from 68.3\% to 89.4\%, compared with 93.8\% when all four heads are patched (means of two map fits). Thus, a steered pattern partially reproduces the steering effect, while other components of the state also affect whether the patch succeeds.

In Ouro, steered patterns produce 248/256 correct answers, close to 251/256 under full steering. Both the unsteered run and the steered run patched with unsteered patterns score zero. Patches using only values, patterns from an unrelated graph, patterns from the wrong loop, or comparison heads also score zero.

\begin{figure}[!htbp]
\centering
\includegraphics[width=\linewidth]{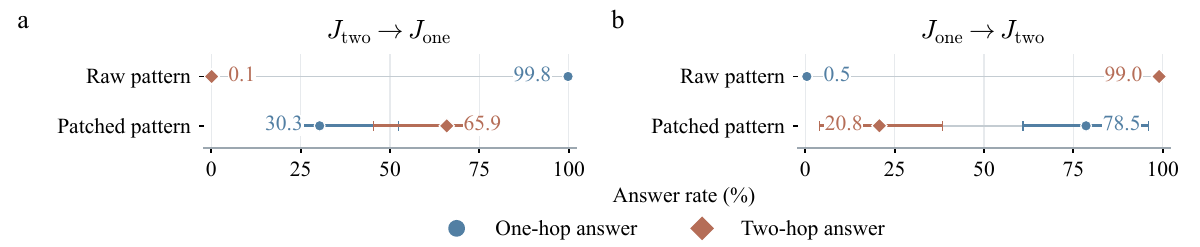}
\caption{Pattern patching between one-hop and two-hop runs. Arrows indicate the patch direction. ``Raw pattern'' uses the run's own attention pattern, while ``Patched pattern'' uses the second-layer pattern from the other run. Points show means across backbones, and horizontal bars show their ranges.}
\label{fig:target-pattern-patching}\label{fig:routing-swap}
\end{figure}

\subsection{Attention Routing Switches the Next Target}
\label{sec:target-routing}
We next ask whether attention patterns determine which of the two controlled targets the model predicts. To test this, we run $F(J_{\mathrm{one}}(h_6))$ and $F(J_{\mathrm{two}}(h_6))$ on the same input. The two runs share the same state before $J$ and the same frozen block. Let $(\alpha_{\mathrm{one}},V_{\mathrm{one}})$ and $(\alpha_{\mathrm{two}},V_{\mathrm{two}})$ be their second-layer attention patterns and values. We patch patterns across all heads and token positions:
\begin{equation}
 z_{\mathrm{two}\rightarrow\mathrm{one}}=\alpha_{\mathrm{two}}V_{\mathrm{one}},\qquad
 z_{\mathrm{one}\rightarrow\mathrm{two}}=\alpha_{\mathrm{one}}V_{\mathrm{two}}.
\end{equation}
Each patched run keeps its own values but takes the attention pattern from the run using the other map.

Patching second-layer attention patterns switches the predicted target in both directions (Figure~\ref{fig:target-pattern-patching}). Patching two-hop patterns into the one-hop run raises the frequency of two-hop answers from 0.1\% to 65.9\%. The reverse patch raises the frequency of one-hop answers from 0.5\% to 78.5\%. Both effects appear in all three selected backbones.

Restoring head outputs provides a complementary test. After replacing queries causes errors, restoring selected outputs repairs 74.0\% of eligible D8L6 errors and all 249 eligible Ouro errors. Restoring comparison heads repairs 0.5\% and none, respectively. Together, these interventions identify attention routing as a causal pathway through which the entering state controls the next target. We next examine how backbone training affects this control.

\section{Steering Depends on the Backbone's Learned Algorithms}
\label{sec:generalization}\label{sec:supervision-boundary}
The preceding experiments show that attention patterns can switch between controlled targets. The three graph backbones were selected because their readouts showed both one-hop and two-hop progression. We now ask whether $J$ and the frozen $F$ can implement a transition that the backbone has not learned. To study this question, we change the backbone supervision before fitting $J$.

We train five pairs of ten-node D8L8 models. Within each pair, the models have identical initializations, inputs, architectures, and update budgets. Every query requests eight hops. Final-only training supervises loop 8 and leaves intermediate predictions unconstrained. Stepwise training adds losses for $f_G^t(s)$ at loops 1--7 to encourage one-hop progression. Writing $R(h_t)$ for the readout distribution and $\ell_t=\operatorname{CE}(R(h_t),f_G^t(s))$, the objectives are
\begin{equation}
\mathcal{L}_{\mathrm{final}}=\ell_8,\qquad
\mathcal{L}_{\mathrm{stepwise}}=\ell_8+\frac{1}{7}\sum_{t=1}^{7}\ell_t.
\end{equation}

We then freeze each backbone and fit separate maps for staying at the current node, advancing one hop, and advancing two hops. Each map acts on $h_8$ and is followed by one frozen loop. Both supervision regimes use the same map family and fitting protocol.

\begin{figure}[!htbp]
\centering
\includegraphics[width=\linewidth]{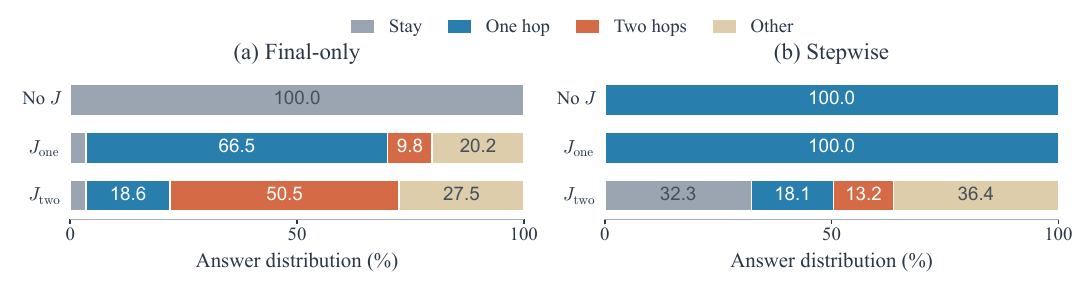}
\caption{Matched graph supervision. Colors mark answers relative to $u=f_G^8(s)$. Bars average two map fits per backbone and five paired backbone seeds.}
\label{fig:matched-supervision}
\end{figure}

Supervision changes both unsteered behavior and steering accuracy (Figure~\ref{fig:matched-supervision}). Unsteered final-only models retain the answer, whereas stepwise models advance one hop. With $J$, one-hop accuracy is 100\% for stepwise models and 66.5\% for final-only models. The ordering reverses for two hops: accuracy is 50.5\% for final-only models and 13.2\% for stepwise models. Final-only models have higher two-hop accuracy in all five pairs. Thus, the transitions reached by the same map family depend on backbone supervision. Appendix~\ref{app:matched-supervision} reports the training protocol and paired comparisons.

We also compare supervision strategies on Ouro antonym cancellation. Each step removes the leftmost adjacent antonym pair. The answer is the pair removed at the requested step. Both backbones are trained for four loops on deletions 1--4. Stepwise training requests four deletions and supervises the current deletion at each loop. Final-only training requests a particular deletion depth and supervises its answer at loop 4.

\begin{figure}[!htbp]
\centering
\includegraphics[width=.90\linewidth]{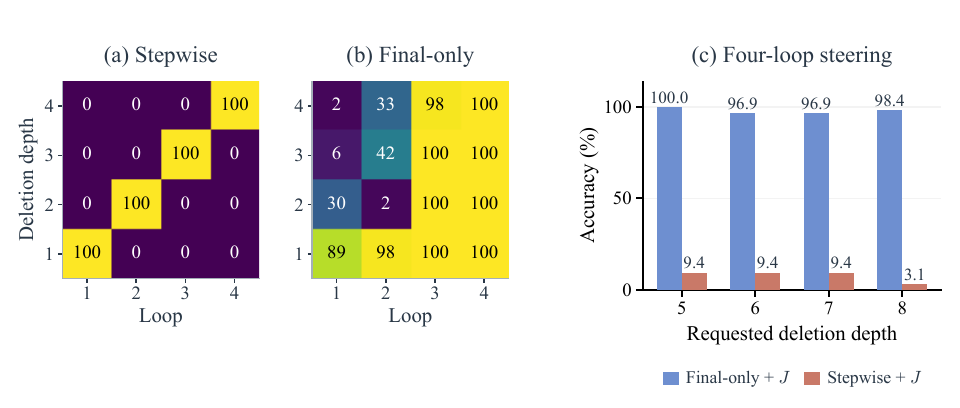}
\caption{Ouro antonym cancellation. (a,b) Readout accuracy (\%) as a function of loop and deletion depth. (c) Accuracy using four loops and a fitted map.}
\label{fig:ouro-native-readouts}\label{fig:steering-boundaries}
\end{figure}

The two strategies produce different readout sequences. The stepwise model predicts one deletion per loop. The final-only model answers 255/256 requests correctly by loop 3 (Figure~\ref{fig:ouro-native-readouts}a,b).

We freeze each backbone and fit a shared rank-128 residual affine map on deletions 1--8 with four loops. The map acts after each loop's RMSNorm, including before the final output head. Depths 5--8 are included in map training. On held-out sequences at these depths, accuracy is 98.0\% for final-only plus $J$ and 7.8\% for stepwise plus $J$. Both unsteered models score zero (Figure~\ref{fig:steering-boundaries}c). Appendix~\ref{app:supervision} provides the Ouro task, training, and evaluation details.

\FloatBarrier

These results show that supervision changes both the unsteered readouts and the targets reached by the fitted maps. We interpret this dependence as evidence that steering relies on the algorithms learned by the backbone.

\FloatBarrier
\section{Conclusion}
Looped Transformers reuse the same parameters across recurrent depth, but our results show that they need not reuse them in the same way. The computation performed by a shared block depends on the hidden state with which it is entered. By modifying only this state, we can redirect the same frozen block toward different semantic transitions. Attention interventions further show that this control is implemented through routing: patching attention patterns can recover steering effects and switch the predicted target while preserving the receiving run's values. Matched supervision experiments show that which transitions are controllable depends on what the backbone has learned, supporting the view that steering redirects computations implemented by the frozen model rather than supplying an arbitrary new computation.

These findings suggest a view of recurrent computation in which the shared Transformer block acts as a reusable executor, while the evolving hidden state determines how that executor is routed across loops. Weight sharing therefore does not imply a fixed ``one loop = one algorithmic step'' rule. At the same time, this reuse has a finite operating regime: controllers that work over one or two loops degrade under longer composition. Successful long-horizon recurrent reasoning therefore requires not only reusable computation, but also states that remain compatible with repeatedly routing that computation.

\subsection*{AI Use Statement}
In this work, we used ChatGPT and Codex primarily to assist with code implementation and debugging, literature search, and manuscript refinement. During writing, these tools helped improve wording, clarity, and readability; the authors determined the scientific claims and conclusions presented in the manuscript. All AI-assisted work was manually reviewed. We checked code for correctness, verified references, and reviewed the text for accuracy and consistency with the experimental results. The authors take full responsibility for the final content of this work, including all text, claims, and artifacts produced with the aid of generative AI.

\subsection*{Ethics Statement}
The experiments study recurrent computation using synthetic graph, parity, and rule-based language tasks. They do not involve human participants. The attention interventions are used to analyze model computation; the experiments do not evaluate the safety of deploying steered models in downstream applications.

\subsection*{Reproducibility Statement}
Reproduction materials are available in the repository linked below the abstract. Please refer to its README for setup and reproduction instructions.

\bibliography{references}

@article{blayney2026mechanistic,
  title={A Mechanistic Analysis of Looped Reasoning Language Models},
  author={Blayney, Hugh and Arroyo, {\'A}lvaro and Obando-Ceron, Johan and Castro, Pablo Samuel and Courville, Aaron and Bronstein, Michael M. and Dong, Xiaowen},
  journal={arXiv preprint arXiv:2604.11791},
  year={2026},
  url={https://arxiv.org/abs/2604.11791}
}

@inproceedings{dehghani2019universal,
  title={Universal {Transformers}},
  author={Dehghani, Mostafa and Gouws, Stephan and Vinyals, Oriol and Uszkoreit, Jakob and Kaiser, {\L}ukasz},
  booktitle={International Conference on Learning Representations},
  year={2019},
  url={https://arxiv.org/abs/1807.03819}
}

@inproceedings{fan2025looped,
  title={Looped {Transformers} for Length Generalization},
  author={Fan, Ying and Du, Yilun and Ramchandran, Kannan and Lee, Kangwook},
  booktitle={International Conference on Learning Representations},
  year={2025},
  url={https://proceedings.iclr.cc/paper_files/paper/2025/hash/25cc3adf8c85f7c70989cb8a97a691a7-Abstract-Conference.html}
}

@article{fan2026lotus,
  title={Bridging the Gap Between Latent and Explicit Reasoning with Looped {Transformers}},
  author={Fan, Ying and Svete, Anej and Lee, Kangwook},
  journal={arXiv preprint arXiv:2606.31779},
  year={2026},
  url={https://arxiv.org/abs/2606.31779}
}

@inproceedings{gatmiry2024gradient,
  title={Can Looped {Transformers} Learn to Implement Multi-step Gradient Descent for In-context Learning?},
  author={Gatmiry, Khashayar and Saunshi, Nikunj and Reddi, Sashank J. and Jegelka, Stefanie and Kumar, Sanjiv},
  booktitle={Proceedings of the 41st International Conference on Machine Learning},
  volume={235},
  pages={15130--15152},
  year={2024},
  publisher={PMLR},
  url={https://proceedings.mlr.press/v235/gatmiry24b.html}
}

@inproceedings{geiping2025scaling,
  title={Scaling up Test-Time Compute with Latent Reasoning: A Recurrent Depth Approach},
  author={Geiping, Jonas and McLeish, Sean and Jain, Neel and Kirchenbauer, John and Singh, Siddharth and Bartoldson, Brian R. and Kailkhura, Bhavya and Bhatele, Abhinav and Goldstein, Tom},
  booktitle={Advances in Neural Information Processing Systems},
  volume={38},
  url={https://proceedings.neurips.cc/paper_files/paper/2025/hash/3b01972cf31e6fa0fe29e4b8b5c2a0a1-Abstract-Conference.html},
  year={2025},
  pages={41340--41391},
  doi={10.52202/085713-1380}
}

@inproceedings{giannou2023loops,
  title={Looped {Transformers} as Programmable Computers},
  author={Giannou, Angeliki and Rajput, Shashank and Sohn, Jy-Yong and Lee, Kangwook and Lee, Jason D. and Papailiopoulos, Dimitris},
  booktitle={Proceedings of the 40th International Conference on Machine Learning},
  volume={202},
  pages={11398--11442},
  series={Proceedings of Machine Learning Research},
  publisher={PMLR},
  year={2023},
  url={https://proceedings.mlr.press/v202/giannou23a.html}
}

@inproceedings{lu2025probing,
  title={Latent Chain-of-Thought? Decoding the Depth-Recurrent Transformer},
  author={Lu, Wenquan and Yang, Yuechuan and Lee, Kyle and Li, Yanshu and Liu, Enqi},
  year={2025},
  url={https://arxiv.org/abs/2507.02199},
  booktitle={First Workshop on the Application of {LLM} Explainability to Reasoning and Planning at {COLM} 2025}
}

@article{popescu2026adaptive,
  title={Adaptive Depth in Looped {Transformers}: Diagnosing Learned Halting Gates and Trajectory Readouts},
  author={Popescu, Andrei Cristian and S{\'a}ez de Oc{\'a}riz Borde, Haitz and Li{\`o}, Pietro},
  journal={arXiv preprint arXiv:2607.20519},
  year={2026},
  url={https://arxiv.org/abs/2607.20519}
}

@inproceedings{saunshi2025looped,
  title={Reasoning with Latent Thoughts: On the Power of Looped {Transformers}},
  author={Saunshi, Nikunj and Dikkala, Nishanth and Li, Zhiyuan and Kumar, Sanjiv and Reddi, Sashank J.},
  booktitle={International Conference on Learning Representations},
  year={2025},
  url={https://arxiv.org/abs/2502.17416}
}

@article{sharma2026readout,
  title={Dense Supervision Is Not Enough: The Readout Blind Spot in Looped Language Models},
  author={Sharma, Rituraj and Vu, Tu},
  journal={arXiv preprint arXiv:2606.24898},
  year={2026},
  url={https://arxiv.org/abs/2606.24898}
}

@inproceedings{singh2024surgery,
  title={Representation Surgery: Theory and Practice of Affine Steering},
  author={Singh, Shashwat and Ravfogel, Shauli and Herzig, Jonathan and Aharoni, Roee and Cotterell, Ryan and Kumaraguru, Ponnurangam},
  booktitle={Proceedings of the 41st International Conference on Machine Learning},
  series={Proceedings of Machine Learning Research},
  volume={235},
  pages={45663--45680},
  year={2024},
  url={https://proceedings.mlr.press/v235/singh24d.html},
  publisher={PMLR}
}

@article{turner2023activation,
  title={Steering Language Models with Activation Engineering},
  author={Turner, Alexander Matt and Thiergart, Lisa and Leech, Gavin and Udell, David and Vazquez, Juan J. and Mini, Ulisse and MacDiarmid, Monte},
  journal={arXiv preprint arXiv:2308.10248},
  year={2023},
  url={https://arxiv.org/abs/2308.10248}
}

@inproceedings{wang2023ioi,
  title={Interpretability in the Wild: A Circuit for Indirect Object Identification in {GPT}-2 Small},
  author={Wang, Kevin Ro and Variengien, Alexandre and Conmy, Arthur and Shlegeris, Buck and Steinhardt, Jacob},
  booktitle={International Conference on Learning Representations},
  year={2023},
  url={https://arxiv.org/abs/2211.00593}
}

@inproceedings{yang2024learning,
  title={Looped {Transformers} are Better at Learning Learning Algorithms},
  author={Yang, Liu and Lee, Kangwook and Nowak, Robert D. and Papailiopoulos, Dimitris},
  booktitle={International Conference on Learning Representations},
  year={2024},
  url={https://proceedings.iclr.cc/paper_files/paper/2024/hash/b8402301e7f06bdc97a31bfaa653dc32-Abstract-Conference.html}
}

@article{zhang2026recurrence,
  title={When Does Recurrence Become an Algorithm? Convergence Selection in Weight-Tied Looped {Transformers}},
  author={Zhang, Tong and Hu, Junhao and Peng, Yun and Xie, Tao},
  journal={arXiv preprint arXiv:2607.20594},
  year={2026},
  url={https://arxiv.org/abs/2607.20594}
}

@article{zhu2025ouro,
  title={Scaling Latent Reasoning via Looped Language Models},
  author={Zhu, Rui-Jie and Wang, Zixuan and Hua, Kai and Zhang, Tianyu and Li, Ziniu and Que, Haoran and Wei, Boyi and Wen, Zixin and Yin, Fan and Xing, He and Li, Lu and Shi, Jiajun and Ma, Kaijing and Li, Shanda and Kergan, Taylor and Smith, Andrew and Qu, Xingwei and Hui, Mude and Wu, Bohong and Min, Qiyang and Huang, Hongzhi and Zhou, Xun and Ye, Wei and Liu, Jiaheng and Yang, Jian and Shi, Yunfeng and Lin, Chenghua and Zhao, Enduo and Cai, Tianle and Zhang, Ge and Huang, Wenhao and Bengio, Yoshua and Eshraghian, Jason},
  journal={arXiv preprint arXiv:2510.25741},
  year={2025},
  url={https://arxiv.org/abs/2510.25741}
}

@inproceedings{vaswani2017attention,
  title={Attention Is All You Need},
  author={Vaswani, Ashish and Shazeer, Noam and Parmar, Niki and Uszkoreit, Jakob and Jones, Llion and Gomez, Aidan N. and Kaiser, {\L}ukasz and Polosukhin, Illia},
  booktitle={Advances in Neural Information Processing Systems},
  volume={30},
  year={2017},
  url={https://arxiv.org/abs/1706.03762}
}
\bibliographystyle{iclr2027_conference}

\appendix

\clearpage
\section*{Part I: Additional Experiments and Discussion}
This part tests whether maps trained for repeated use remain accurate beyond their training range. We use separate parity and eight-node graph models. These experiments extend the reuse tests in Section~\ref{sec:composition}, where each map was trained for one additional loop.

\section{Depth Extrapolation}
\label{app:depth-extrapolation}\label{sec:depth-boundary}
Section~\ref{sec:composition} reuses maps trained for one additional loop. Here we train maps for repeated use and evaluate them beyond the training range. We study parity and a separate set of twelve eight-node D8L8 backbones.

Parity predicts the sum of binary inputs modulo two after $t=n$ loops \citep{fan2025looped}. Backbone training uses lengths 1--20, and map training uses lengths 20--40. One fixed map acts before loops 2 through $n$. In graph continuation, map training supervises successive nodes through loop 16 using states generated by the steered run. The fixed map acts before every loop. We score the prediction at loop $t$ against $f_G^t(s)$ while keeping the query fixed at depth 8 (Sections~\ref{app:parity} and~\ref{sec:boundary}).

\begin{figure}[!htbp]
\centering
\setcounter{subfigure}{0}%
\begin{subfigure}[t]{.495\linewidth}
\includegraphics[width=\linewidth]{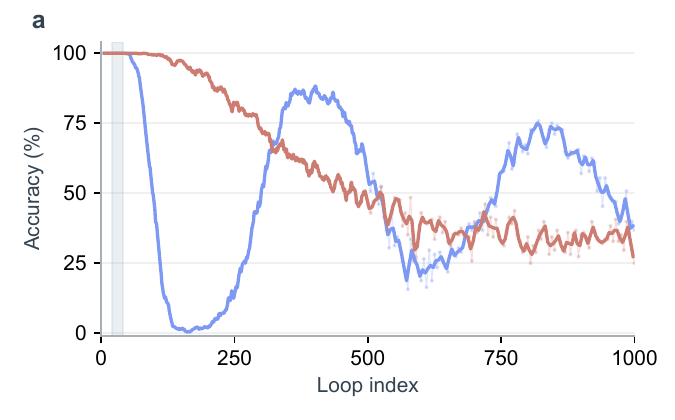}
\phantomcaption\label{fig:v64-parity}
\end{subfigure}\hfill
\begin{subfigure}[t]{.495\linewidth}
\includegraphics[width=\linewidth]{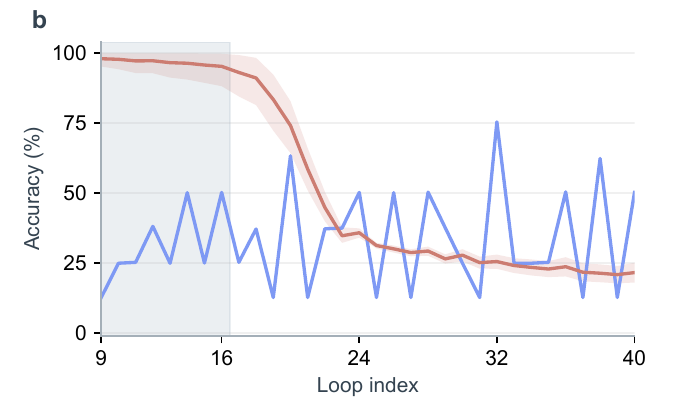}
\phantomcaption\label{fig:v64-graph}
\end{subfigure}
\par\nointerlineskip
\includegraphics[width=\linewidth]{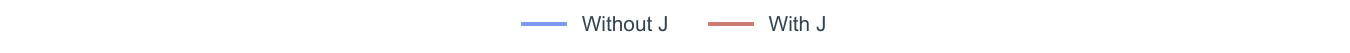}

\caption{Execution beyond map-training ranges. (a) Parity seed 2 at $t=n$: dark curves are smoothed; faint points are individual measurements. (b) Graph continuation: lines average twelve backbones and two maps each; shading shows 95\% bootstrap intervals over backbone means. Gray regions mark map-training ranges.}
\label{fig:boundary}
\end{figure}

For the parity seed shown, steering improves accuracy over lengths 101--500 from 45.05\% to 73.73\%. At measured lengths 505--1000, accuracy is instead 47.55\% without $J$ and 36.40\% with $J$ (Figure~\ref{fig:boundary}a). The improvement therefore does not persist over the full evaluated range.

In graph continuation, all twelve backbones lose accuracy beyond the map-training range. On an independent held-out cycle subset, accuracy falls from 97.42\% at loops 9--16 to 35.11\% at loops 17--32. Figure~\ref{fig:boundary}b shows a separate, broader graph sample. We have not verified that its graph identities are excluded from training. These results show that accurate continuation within the map-training range does not ensure accuracy at greater depth.

\subsection{Parity}
\label{app:parity}\label{sec:parity}

\subsubsection{Task Definition and Input Format}
We use binary parity to compare the requested loop with the loops at which the model predicts the correct answer.

A parity input contains binary tokens, a separator, and optional padding. Token IDs 0 and 1 represent bits, 2 marks the separator, and 3 marks padding. For example, \texttt{1 0 1 1 2} has target 1 because it contains three ones. The sequence \texttt{1 1 0 0 2} has target 0.

We supervise the answer at the separator and any padding tokens that follow it. With padding, the first example becomes \texttt{1 0 1 1 2 3 3}, and the target from the separator onward is \texttt{1 3 3}. Input-bit positions are excluded from the answer loss. The semantic answer is the parity bit, but the implementation scores the complete answer region, including padding.

\subsubsection{Model and Steering Configuration}

For an input of length $n$, we supervise the answer after $n$ loops. Backbone training uses lengths 1--20. We then freeze the backbone and train the map on lengths 20--40. Both stages use cross-entropy on the answer region. The same rank-48 diagonal-plus-low-rank affine map acts before loops 2 through $n$. No intermediate loop receives a partial-parity target. Evaluation at longer lengths keeps the map fixed and increases both the input length and the requested loop count.

The task and length-dependent loop schedule follow \citet{fan2025looped}. We use causal attention without positional encodings. Embeddings enter at the first loop only. Each of the three independently trained backbones has one fitted map.

The shared parity block has one layer, hidden dimension 256, 64 attention heads, and MLP dimension 1,024. Backbone training uses 100,001 AdamW updates, batch size 64, learning rate $10^{-4}$, weight decay 0.01, and gradient clipping at 1.0. Training is in FP32. The learning rate is held constant through the initial curriculum and then follows cosine decay. Validation at length 20 is evaluated every 1,000 updates.

The map starts as the identity and uses three training stages: lengths 20--24 for 256 updates with batch size 64, lengths 20--32 for 512 updates with batch size 48, and lengths 20--40 for 1,024 updates with batch size 32. The stages total 1,792 updates and 73,728 examples. They use AdamW with zero weight decay, peak learning rate $2\times10^{-5}$, 1,024 warmup updates, and cosine decay to one tenth of the peak. The diagonal term uses one tenth of the learning rate of the other map parameters. Gradients are clipped at 1.0, and evaluation uses the final map. Lengths 20--40 are therefore all covered, but their total training frequencies are unequal.

\subsubsection{Evaluation and Readout-Timing Analysis}
\label{app:parity-evaluation}
We evaluate the fitted maps at longer lengths and then measure where accurate readouts occur relative to the requested exit.

\paragraph{Dense evaluation from length 500 to 1000.}
The extended curve uses the seed-2 backbone at update 99,000 and its map trained on lengths 20--40. We evaluate 101 lengths, $500,505,\ldots,1000$, with 128 binary inputs per length. The unsteered and steered runs use the same inputs, generated with seed $2026092401+1009n$. We retain all 12,928 sequences. Embeddings enter only at loop 1, and $J$ acts before loops 2 through $n$.

The original evaluation through length 500 uses 64 examples per length and a 10-length moving mean in Figure~\ref{fig:boundary}a. The extended evaluation starts at length 500 and samples every fifth length with 128 paired examples. Its dark curves average adjacent measured points; faint points show individual measurements. Gray shading marks map-training lengths 20--40. Over the 100 sampled lengths strictly above 500, unsteered and steered accuracy average 47.55\% and 36.40\%, respectively. Each length receives equal weight.

\paragraph{Heatmaps and readout offset.}
The parity heatmaps use 256 examples for each input length and loop count up to 60. Insets display lengths and loops 54--60 without interpolation; dots mark the unique best loop among $t=n-1,n,n+1$ within the measured grid. The dotted horizontal line marks the longest map-training length, $n=40$.

For each seed, the heatmaps, accuracy curves, and timing measurements use the same saved map and intervention schedule. We estimate the positions of the accuracy bands from the period-four Fourier component along the offset between loop number and input length. We unwrap its phase over input length to obtain a continuous estimate of band position. We then fit a line over lengths 41--490.

We compute confidence intervals from 5,000 moving-block bootstrap samples of the fit residuals, with block length 20. This estimate describes the visible accuracy bands; it does not establish an oscillator in the hidden state. We report drift, residual error, and accuracy at the requested loop for all three seeds below.

For the displayed seed 2, the fitted slope of the estimated readout position changes from 0.9904 to 0.9982. Near length 60, steering shifts the best nearby integer readout from $t=n-1$ to $t=n$. These estimates describe readout timing over the measured range.

\subsubsection{Readout Timing across Backbones}
\label{app:parity-results}
We apply the same timing analysis to all three backbones to compare accuracy gains with changes in drift.

\begin{figure}[!htbp]
\centering
\includegraphics[width=\linewidth]{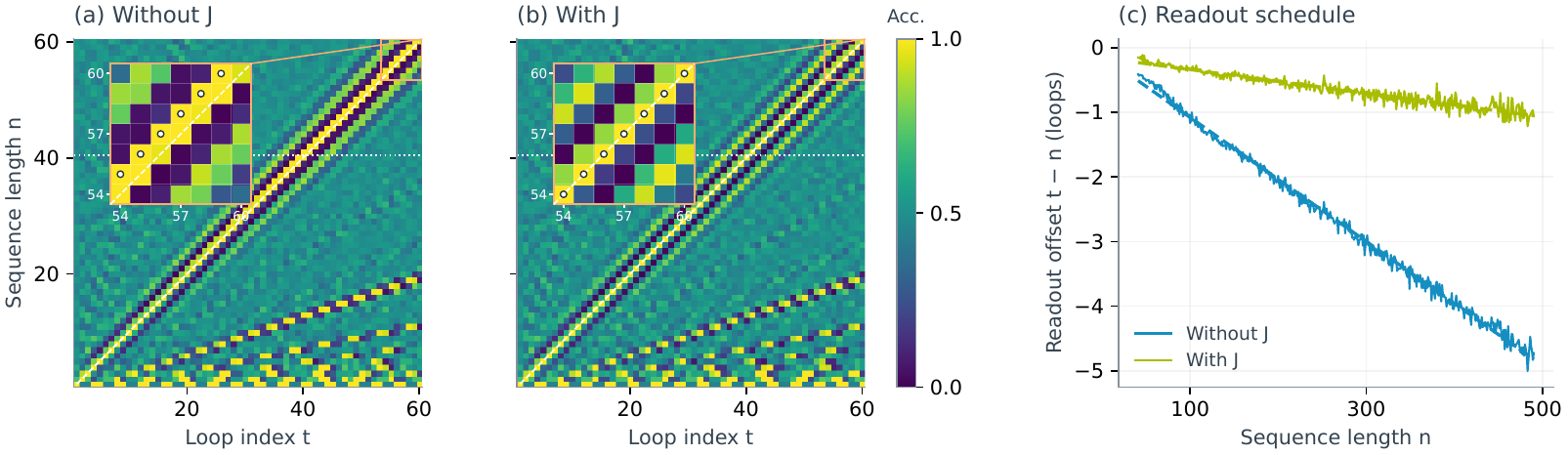}
\caption{Parity readout timing for seed 2. (a,b) Accuracy by input length $n$ and loop $t$, without and with $J$; insets enlarge lengths 54--60. The diagonal marks $t=n$; dots mark the best nearby integer loop. (c) Readout offset $t-n$ and fitted trends.}
\label{fig:parity}\label{fig:parity-phase}
\end{figure}

Changes in readout timing help explain the initial accuracy improvement for seed 2. Without steering, accurate predictions drift earlier than $t=n$. Steering moves them closer to the requested loop (Figure~\ref{fig:parity}). Over lengths 41--490, absolute drift falls from 0.96 to 0.18 loops per 100 tokens. This effect differs across seeds and does not explain the full accuracy curve through length 1000. Section~\ref{app:parity-evaluation} defines the estimator; the table below reports all three backbones.

\paragraph{Long-range accuracy and readout timing.}
\begin{center}\begin{minipage}{\linewidth}\centering
\centering
\captionsetup{hypcap=false}
\captionof{table}{Readout-schedule drift and longer-range accuracy with each fixed steering map. Drift is measured in loops per 100 tokens; MAE is the mean absolute deviation from the fitted phase line. Accuracy averages $t=n$ over lengths 101--500.}
\begin{tabular}{@{}lrrr@{}}
\toprule
Seed & Drift: native $\to J$ & MAE: native $\to J$ & Accuracy: native $\to J$ (\%) \\
\midrule
0 & $+0.573\to-0.327$ & $0.065\to0.069$ & $35.27\to55.71$ \\
1 & $+1.312\to+1.731$ & $0.174\to0.743$ & $44.03\to45.24$ \\
2 & $-0.961\to-0.183$ & $0.066\to0.052$ & $45.05\to73.73$ \\
\bottomrule
\end{tabular}
\end{minipage}\end{center}

Seed 1 shows that higher accuracy at the requested loop need not imply better timing alignment. Its mean accuracy improves slightly with $J$, but both drift and residual error increase. A single linear trend also fits its later readouts poorly.

\subsection{Graph Continuation}
\label{sec:boundary}
We next test whether maps trained for repeated use sustain graph walks beyond the training range. These eight-node D8L8 backbones are separate from the ten-node D8L8 models in Section~\ref{sec:underdetermined} and the D8L6 mechanism models.

\subsubsection{Task Definition and Input Format}
This experiment uses the graph-token format in Appendix~\ref{app:graph-task}, with eight nodes and 29 input tokens. The query continues to specify depth 8 even when execution extends beyond eight loops. At loop $t$, the continuation target is the node reached after $t$ graph edges from the original start. The input is not rewritten with a new requested depth at each loop.

For example, on the cycle $0\to1\to\cdots\to7\to0$ with start 0, the loop-8 target is 0, the loop-9 target is 1, and the loop-10 target is 2.

\subsubsection{Model and Steering Configuration}

The continuation study uses twelve eight-node D8L8 backbones with seeds 100--111. Each repeats a shared two-layer block with hidden dimension 256 and four attention heads per layer. Each reaches perfect endpoint accuracy at the supervised eighth loop. We fit two rank-48 diagonal-plus-low-rank maps per frozen backbone, using cross-entropy on the continuation targets.

Backbone training supervises the requested answer at loop 8. Map training instead supervises successive nodes at loops 1--16 and uses states generated by the steered run. The map acts before every loop, including loop 1 and evaluation loops beyond 16. This schedule differs from the single additional loop used for D8L6 target selection.

Backbone training uses 20,000 AdamW updates, batch size 512, learning rate $3\times10^{-4}$, weight decay 0.3, and 500 warmup updates. We use the final checkpoint. Both the 512-graph selection pool and the 512-graph final test pool are excluded from training.

Each map is trained for 8,000 AdamW updates with batch size 128, learning rate $10^{-4}$, zero weight decay, and gradient clipping at 1.0. Training excludes both graph pools. Validation runs every 400 updates on the selection pool, and we retain the checkpoint with the lowest mean successor cross-entropy over loops 1--16.

\subsubsection{Evaluation Protocol}

We keep the maps fixed and evaluate them beyond loop 16. The broad evaluation sample contains 8,192 unique graphs drawn uniformly without replacement from all 40,320 eight-node permutations, using seed 2026092402. It includes 993 single cycles, as well as graphs with shorter cycles and fixed points. We retain all eight starting nodes and all predictions. Evaluation uses the twelve original checkpoints and their two saved maps, with steering before every loop from loop 1 onward.

We store counts by graph, loop, backbone, and map. We average over starting nodes and graphs, then over maps and backbones. The plotted 95\% interval uses 10,000 bootstrap samples of the twelve backbone means. Gray shading in Figure~\ref{fig:boundary}b marks loops 9--16, the displayed part of the map-training range. The map remains active after loop 16.

We have not checked this broad sample against the historical training exclusions. We therefore also report a separate evaluation on graphs excluded from training.

\subsubsection{Variation across Backbones and Evaluation Sets}
Both populations show lower accuracy beyond the fitted loops, although they average over different graph structures.

Across the twelve backbones, steered accuracy ranges from 75.01\% to 100.00\% over loops 9--16 and from 34.73\% to 54.56\% over loops 17--32. Each value averages the two fixed maps. Every backbone has lower mean accuracy beyond the map-fitting range.

\paragraph{A separate training-excluded evaluation.}
The separate evaluation uses 512 graphs excluded from backbone training, map training, and protocol selection. On its 78 single-cycle graphs, steered accuracy is 97.42\% over loops 9--16 and 35.11\% over loops 17--32. Unsteered accuracy is 12.5\% in both ranges. All twelve backbones lose accuracy beyond the map-training range.

We select this subset by graph structure, without filtering on predictions, and score each loop. Its results are separate from the broad-sample curve. On an eight-node cycle, retaining one endpoint produces a correct answer once every eight loops. The resulting 12.5\% accuracy does not require the model to continue the graph walk.

\subsection{Discussion}
Training a map for repeated use does not ensure that it works at greater depth. Parity steering improves accuracy over an intermediate length range, but its timing effect varies across backbones. The displayed map reduces mean accuracy at measured lengths above 500. In graph continuation, all twelve backbones lose accuracy beyond the training range despite map training on states generated by the steered run.

\FloatBarrier

\section*{Part II: Details and Supporting Results for the Main Text}
This part provides the settings, evaluation protocols, and supporting results for Sections~\ref{sec:underdetermined}--\ref{sec:generalization}. Appendix~\ref{app:all-native-trajectories} gives the complete native readouts. Appendices~\ref{app:main-config} and~\ref{app:parallel} document state control and attention interventions. Appendices~\ref{app:matched-supervision} and~\ref{app:supervision} give the supervision comparisons, followed by a consolidated training summary in Appendix~\ref{app:training-summary}.

\section{Native Readout Trajectories}
\label{app:all-native-trajectories}
The main text shows four native readout trajectories. Here we report all twelve D8L8 seeds (0--11) and five D8L6 seeds (3--7), including the main-text examples. Each model is evaluated on 512 held-out ten-node cycles and all ten starting nodes. We do not filter examples by prediction correctness.

\begin{figure}[!htbp]
\centering
\includegraphics[width=.93\linewidth]{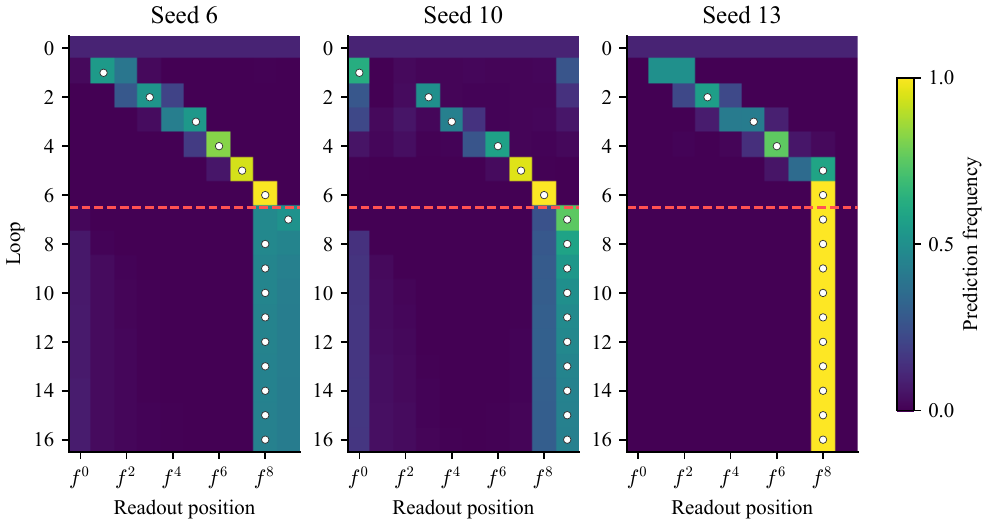}
\caption{Native readouts of the D8L6 backbones used in the attention experiments.}
\label{fig:selected-mechanism-readouts}
\end{figure}

All heatmaps use the visual conventions of Figure~\ref{fig:trajectory}. Color shows prediction frequency, red dashed lines mark the training loop budget, and white dots mark the unique most frequent prediction. Ties are unmarked. Positions on ten-node cycles measure progress modulo ten. Seed 6 is the original D8L6 mechanism model. Figure~\ref{fig:selected-mechanism-readouts} shows the additional selected backbones.

\begin{figure}[!htbp]
\centering
\includegraphics[width=\linewidth]{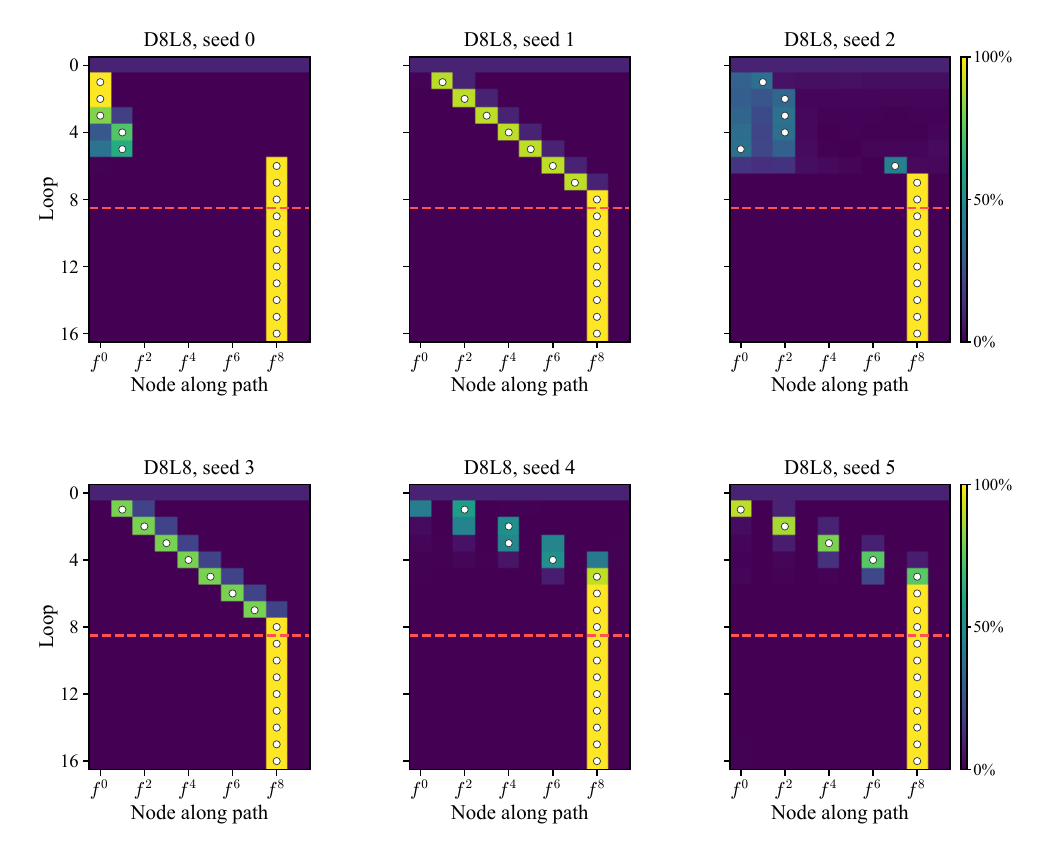}
\caption{Native D8L8 readouts for seeds 0--5.}
\label{fig:all-trajectory-a}
\end{figure}
\clearpage

\begin{figure}[!htbp]
\centering
\includegraphics[width=\linewidth]{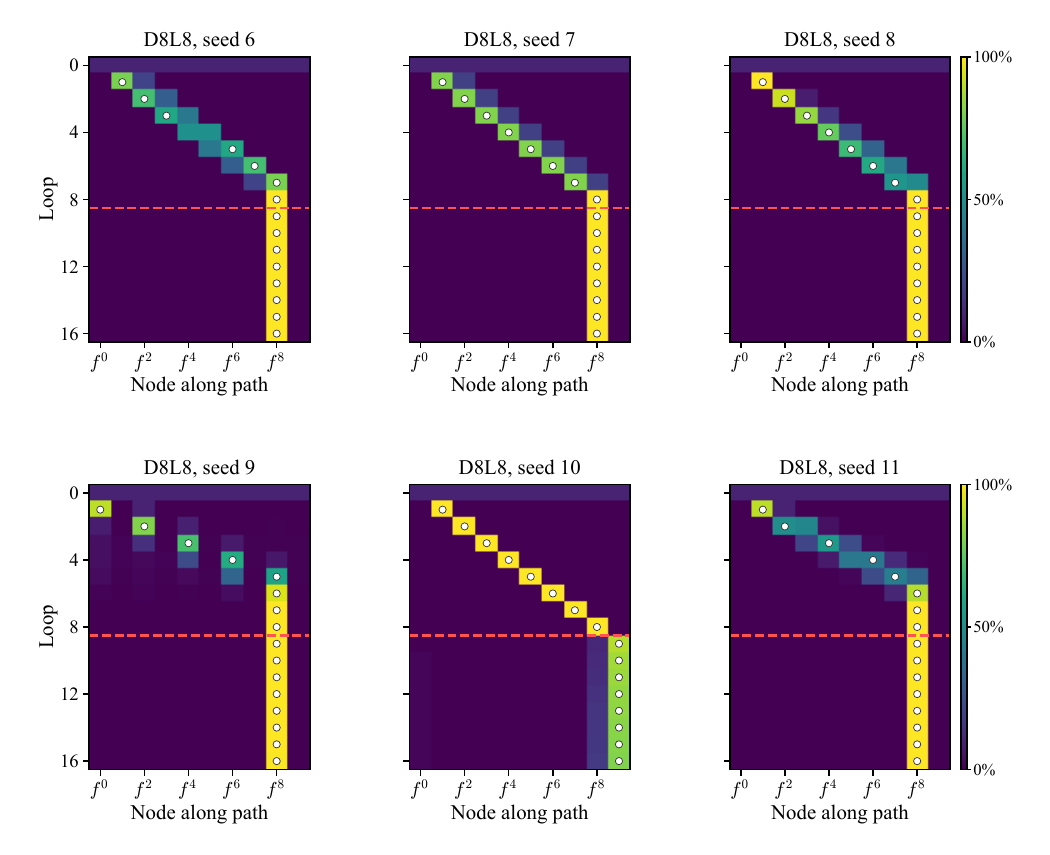}
\caption{Native D8L8 readouts for seeds 6--11.}
\label{fig:all-trajectory-b}
\end{figure}
\clearpage

\begin{figure}[!htbp]
\centering
\includegraphics[width=\linewidth]{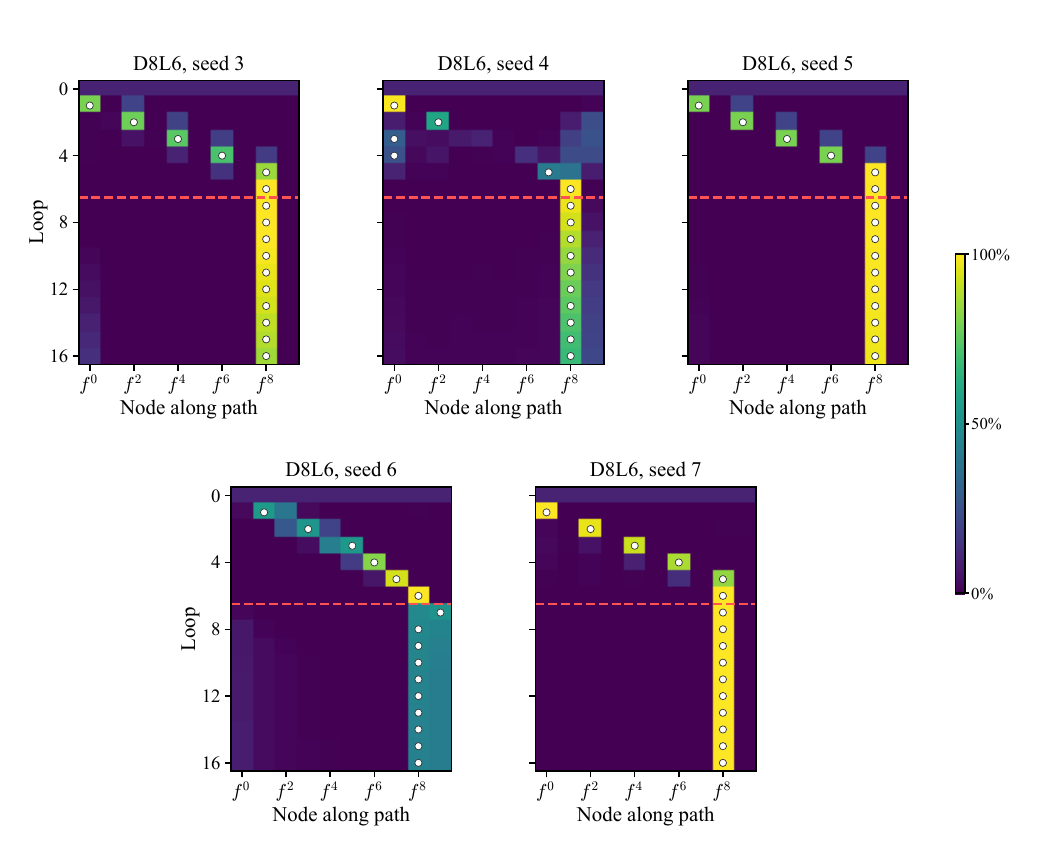}
\caption{Native D8L6 readouts for seeds 3--7.}
\label{fig:all-trajectory-c}
\end{figure}
\clearpage
\subsection{Readout Summary}
\label{app:trajectory-summary}
We summarize the heatmaps through differences between consecutive model predictions.

We use the saved readout frequencies from 512 held-out ten-node cycles and all ten starts per backbone. Let $\hat d_t$ be the unique most frequent decoded path position across these 5,120 examples at loop $t$. We compute $\Delta_t=(\hat d_t-\hat d_{t-1})\bmod 10$ only when both modes are unique. All backbones have a tie at loop 0, so we omit the first increment. D8L8 seed 6 also has a tie at loop 4. We do not break ties arbitrarily.

These increments compare modes across the full evaluation set. They do not describe the distribution of transitions for individual examples. A zero increment, labeled ``Hold,'' means zero decoded displacement modulo ten. It does not mean that the hidden state performs no computation.

\begin{center}\begin{minipage}{\linewidth}\centering
\captionsetup{hypcap=false}
\captionof{table}{The twelve D8L8 backbones and five D8L6 backbones. Model readout increments through the final supervised loop; -- marks an undefined increment due to a tie.}
\begin{tabular}{lll}\toprule
Configuration & Seed & Increments to loops 1 through $L$\\\midrule
D8L8 & 0 & --, 0, 0, 1, 0, 7, 0, 0 \\
D8L8 & 1 & --, 1, 1, 1, 1, 1, 1, 1 \\
D8L8 & 2 & --, 1, 0, 0, 8, 7, 1, 0 \\
D8L8 & 3 & --, 1, 1, 1, 1, 1, 1, 1 \\
D8L8 & 4 & --, 2, 0, 2, 2, 0, 0, 0 \\
D8L8 & 5 & --, 2, 2, 2, 2, 0, 0, 0 \\
D8L8 & 6 & --, 1, 1, --, --, 1, 1, 0 \\
D8L8 & 7 & --, 1, 1, 1, 1, 1, 1, 1 \\
D8L8 & 8 & --, 1, 1, 1, 1, 1, 1, 1 \\
D8L8 & 9 & --, 2, 2, 2, 2, 0, 0, 0 \\
D8L8 & 10 & --, 1, 1, 1, 1, 1, 1, 1 \\
D8L8 & 11 & --, 1, 2, 2, 1, 1, 0, 0 \\
D8L6 & 3 & --, 2, 2, 2, 2, 0 \\
D8L6 & 4 & --, 2, 8, 0, 7, 1 \\
D8L6 & 5 & --, 2, 2, 2, 2, 0 \\
D8L6 & 6 & --, 2, 2, 1, 1, 1 \\
D8L6 & 7 & --, 2, 2, 2, 2, 0 \\
\bottomrule\end{tabular}\end{minipage}\end{center}

\clearpage

\section{Graph Walk}
\label{app:main-config}
This appendix supports Sections~\ref{sec:underdetermined}--\ref{sec:hypothesis}. It defines the graph task and documents the configurations, evaluation protocols, and detailed results for the control and attention experiments reported in the main text.

\subsection{Task Definition and Input Format}
\label{app:graph-task}

The task predicts the node reached after a requested number of edges from a specified start. Each node has a discrete token ID. The input consists of a beginning token, one three-token record per directed edge, and a query:
\begin{quote}
\texttt{BOS [EDGE source destination] ... QUERY start DEPTH[k] ANSWER}
\end{quote}
The brackets group records for display; they are not input tokens. \texttt{BOS}, \texttt{EDGE}, \texttt{QUERY}, and \texttt{ANSWER} are dedicated tokens, as is each supported depth. The model predicts a node from the hidden state at \texttt{ANSWER}. A ten-node input has 35 tokens. The graph is supplied in the input, rather than stored as a fixed set of edges in the model.

For example, consider the ten-node cycle $0\to1\to\cdots\to9\to0$. Its edge records are \texttt{EDGE 0 1}, \texttt{EDGE 1 2}, and so on through \texttt{EDGE 9 0}. The query \texttt{QUERY 0 DEPTH[8] ANSWER} has target node 8. Changing the start to 3 gives target node 1. These examples illustrate the encoding; they are not evaluation samples.

Backbone training samples requested path lengths 1--8. Cross-entropy supervises the requested node at the final training loop: loop 8 for D8L8 and loop 6 for D8L6. Earlier readouts receive no task loss.

For target selection, the request remains fixed at eight steps. Both maps start from the same D8L6 state after loop 6 and are followed by one frozen loop. The one-hop target is the successor of the eight-step answer; the two-hop target is its second successor. In the cycle example starting at 0, the targets are 9 and 0, respectively. No answer node is appended to the input.

The attention interventions use this same input format and target rule. They replace internal attention quantities in already trained runs; they do not introduce a new training target. Their paired inputs and scoring rules are specified in Appendix~\ref{app:n10-mechanism}.

\subsection{Model and Steering Configuration}
\label{app:target-setup}
With the task fixed, we specify the frozen backbone and the maps trained to control its next loop.

\paragraph{Backbone and intervention.}
Target selection uses the ten-node D8L6 model trained with seed 6, at update 16,000. We evaluate selection-set endpoint accuracy every 1,000 updates and retain the earliest checkpoint with the highest accuracy. The model receives an eight-hop request with answer $u=f_G^8(s)$ and runs for six loops to produce $h_6$.

From this state, we compare one unsteered loop, $F(h_6)$, with $F(J_{\mathrm{one}}(h_6))$ and $F(J_{\mathrm{two}}(h_6))$. The input, backbone weights, and number of additional loops remain fixed. Each map acts on every token before the final frozen loop.

The D8L6 backbone has hidden dimension 256, four attention heads, and an MLP dimension of 1,024 in each of its two shared layers. It uses pre-layer normalization, a layer-normalized readout, and no dropout. Backbone training uses 20,000 updates, batch size 512, learning rate $3\times10^{-4}$, weight decay 0.3, and 500 warmup updates. Requested depths are sampled uniformly from 1--8. Graphs are sampled uniformly from permutations after excluding the locked evaluation and selection graphs.

\paragraph{Map dimensions and training.}
Both maps have the form $J(h)=h(D+AB)+b$, with hidden dimension 256 and rank 48. The diagonal $D$ has 256 learned entries, $A\in\mathbb{R}^{256\times48}$, $B\in\mathbb{R}^{48\times256}$, and $b\in\mathbb{R}^{256}$. Each map therefore has 25,088 trainable parameters.

We fit each target map twice, with seeds 1 and 2. Only the map is trained. Cross-entropy supervises $f_G(u)$ for $J_{\mathrm{one}}$ and $f_G^2(u)$ for $J_{\mathrm{two}}$ after the frozen loop. Each fit uses 8,000 AdamW updates, batch size 128, learning rate $10^{-4}$, zero weight decay, and gradient clipping at 1.0. We validate every 400 updates and retain the earliest checkpoint with the highest validation accuracy. Training excludes the fixed selection and confirmation graphs.

\subsection{Evaluation and Attention Interventions}
\label{app:n10-mechanism}
The trained maps remain fixed in all evaluations below; the experiments differ in their paired inputs, eligibility rules, and patched components.

\paragraph{Native intermediate readouts.}
For Figure~\ref{fig:trajectory}, we request the eight-step answer and evaluate the frozen models at loops 0--16. Loop 0 precedes the first shared block. Each model is evaluated on 512 held-out ten-node cycles with all ten starting nodes, giving 5,120 examples. The figure shows four models from twelve D8L8 runs and the five D8L6 runs. Evaluation extends beyond the six or eight training loops without changing the weights.

\paragraph{Target-selection accuracy.}
Figure~\ref{fig:target-behavior} uses the same 4,110 examples across the unsteered and steered conditions. The endpoint, one-hop, and two-hop labels are distinct, and examples are not filtered by prediction correctness. We classify each answer as the endpoint, its first successor, its second successor, or another node, and average the steered results across the two fits per target. The subsequent mechanism tests retain the same backbone and one-hop maps but use their own paired evaluation inputs.

\paragraph{Head selection and paired inputs.}
The following details describe the original seed-6 evaluation. The three-backbone aggregate and replication cohorts are specified in Appendix~\ref{app:selected-mechanism}. The seed-6 tests reuse the frozen checkpoint and two one-hop maps specified above.

We select head H2 in the second layer on an earlier discovery set of 32 graphs. It has the highest mean eligible pattern-patching accuracy across the two map fits, with ties resolved by head index. H3 is the prespecified comparison head. The confirmation keeps this choice fixed and retains results for all four heads.

For confirmation, sampling seed 2026092404 draws 512 graph pairs without replacement from a reserve of 17,519 unused permutation graphs. All 1,024 identities are excluded from backbone training, map training, earlier mechanism evaluations and their generated variants, and the preceding 1,024-graph evaluation of the main D8L6 backbone. The reserve was constructed by single swaps in earlier held-out graphs.

We draw the two graphs in each pair independently; they need not share most edges. Node slots align between inputs. We patch activations from run 1 into run 2. We evaluate all ten current nodes in run 2 and set the current node in run 1 three positions ahead modulo ten. For each graph, we choose the starting node so that eight graph steps reach the designated current node. Each run executes six loops, applies the map to all tokens, and executes a seventh frozen loop.

\paragraph{Scoring populations.}
These choices produce 5,120 candidate pairs before patching. The pattern-versus-output experiment requires three distinct answers: the original answer in run 2, the answer in run 1, and the answer obtained by following the patched attention pattern in the graph used for run 2. Both unpatched runs must correctly read their current and next nodes. This leaves 3,966 pairs for fit 1 and 3,963 for fit 2; within each fit, both patch types use exactly the same pairs.

For restoration, the alternative-current run must have correct current and next readouts and a distinct target. Recovery is scored among eligible clean transitions broken by query replacement, giving 4,413 and 4,417 cases. Steering-pattern patching uses 4,044 examples with distinct current, one-hop, and two-hop labels, without conditioning on prediction correctness. The earlier target-selection figure uses a separate cohort of 4,110 examples. Each experiment retains its own denominator.

\paragraph{Uncertainty and implementation checks.}
For each map fit, we bootstrap 5,000 samples of the 512 graph pairs. All ten starting nodes of a graph remain together. The two fits measure variation between maps on one backbone, rather than between independently trained backbones.

Direct evaluation, reconstructed attention, an independent output hook, and same-run activation replacement agree on predictions when no effective intervention is made.

\begin{table}[!htbp]
\centering
\caption{D8L6 evaluation populations. Counts are graph--start instances or paired runs, as appropriate. All map fits within a row use the stated graph population; correctness-based eligibility can differ by fit.}
\label{tab:graph-populations}
\begin{tabular}{@{}>{\raggedright\arraybackslash}p{.23\linewidth}>{\raggedright\arraybackslash}p{.20\linewidth}>{\raggedright\arraybackslash}p{.49\linewidth}@{}}\toprule
Experiment & Scored count & Inclusion and intervention\\\midrule
Native trajectories & 5,120 per backbone & 512 ten-node cycles; all starts; no prediction filtering.\\
Target selection & 4,110 & Distinct current, one-hop, and two-hop labels; no prediction filtering; one additional $F$.\\
Pattern vs. output & 3,966 / 3,963 & Distinct semantic answers and correct clean readouts in both runs; selected second-layer head H2.\\
Steering-pattern patching, Figure~\ref{fig:parallel-mechanism}c & 4,044 & Distinct current, one-hop, and two-hop labels; no prediction filtering; all L2 heads at the answer position.\\
Output restoration & 4,413 / 4,417 & Eligible clean transitions broken by query replacement; restore H2 or control H3.\\
Target switching & 4,116 & Separate 512-graph cohort; distinct current, one-hop, and two-hop labels; all second-layer heads and token positions.\\
Original two- and eight-step composition & 3,200 & Same 512 graphs as target switching; $u$ through $f_G^4(u)$ distinct; no prediction filtering.\\
\bottomrule\end{tabular}
\end{table}

\subsection{Supporting Mechanism and Control Results}
\label{app:graph-results}
Using these evaluation rules, we report output restoration, target switching, and controller reuse.

\subsubsection{Restoring Head Outputs Recovers Answers}
\label{sec:restore}
We test whether restoring selected head outputs repairs errors caused by query replacement.

We run the same graph with two different starting nodes. The original run provides the clean answer and head outputs. The alternative run provides replacement queries. We insert these queries into the original run while keeping its keys and values unchanged. This changes where attention reads without replacing the input's value vectors. We then replace the selected heads' disrupted outputs with their clean outputs and let the model continue. For an intervened head, let $Q_{\mathrm{alt}}$ denote the replacement queries and $d_k$ the query/key dimension:
\begin{equation}
 z_{\mathrm{broken}}
 =\operatorname{softmax}\!\left(\frac{Q_{\mathrm{alt}}K_{\mathrm{clean}}^\top}{\sqrt{d_k}}\right)V_{\mathrm{clean}},
 \qquad
 z_{\mathrm{restored}}\leftarrow z_{\mathrm{clean}},
\end{equation}
Here softmax is taken over the visible key positions. If the selected outputs carry information needed for the answer, restoring them should repair the error caused by query replacement. We compare this repair with restoring a control set of heads. Recovery is measured only on examples that were answered correctly before query replacement and incorrectly afterward.

Restoring the selected outputs repairs errors in both models. In the original D8L6 seed-6 evaluation, it repairs 60.1\% of disrupted answers, compared with 0.9\% for the control. In Ouro, it repairs 249 of 249 answers, while the matched control repairs 0. Restoring all affected outputs repairs every disrupted answer in both models. The selected outputs therefore provide information that later computation uses to produce the answer.

\subsubsection{Patching One-Hop and Two-Hop Patterns}
\label{app:target-exchange}
We reuse the frozen seed-6 backbone and both pairs of target-specific maps. Sampling seed 2026092501 draws 512 fresh graphs from the training-excluded reserve, excluding previous evaluation graphs and their generated variants. Evaluating all ten starting nodes yields 5,120 examples. We retain the 4,116 examples with distinct current, one-hop, and two-hop labels, without filtering by prediction correctness. Both directions and both map fits use this same population. This cohort is separate from the 4,110-example target-selection figure and the 4,044-example steering-pattern experiment.

The two runs share the input and $h_6$ but use different maps before the final frozen loop. We patch post-softmax patterns at all token positions and all four heads of layer 2. Each patched run keeps its own values and continues normally.

Same-run replacements preserve predictions. Direct evaluation and the intervention implementation agree on unpatched predictions; checkpoint hashes remain unchanged. Confidence intervals use 5,000 paired bootstrap samples of graph identities, retaining all starting nodes and both map fits together. The two fits quantify map-fit variation on a fixed backbone.

\begin{center}\begin{minipage}{\linewidth}\centering\centering
\captionsetup{hypcap=false}
\captionof{table}{Seed-6 answers matching the target of the run providing the second-layer patterns (percent). Arrows indicate the direction of patching. Confidence intervals describe the two-fit mean, with graphs as bootstrap units.}
\begin{tabular}{@{}lrrrr@{}}\toprule
Direction & Fit 1 & Fit 2 & Mean & 95\% CI\\\midrule
$J_{\mathrm{two}}\!\to\!J_{\mathrm{one}}$ & 44.97 & 45.82 & 45.40 & [44.43, 46.36] \\
$J_{\mathrm{one}}\!\to\!J_{\mathrm{two}}$ & 59.67 & 62.17 & 60.92 & [59.84, 61.97] \\
\bottomrule\end{tabular}\end{minipage}\end{center}

\subsubsection{Composition of Frozen Target-Specific Maps}
\label{app:composition}
We evaluate D8L6 seed 6 and both existing pairs of one-hop and two-hop maps without further training. The first continuation applies a map to $h_6$ and executes the seventh loop. The second applies a map to the resulting full token-state sequence and executes the eighth loop. The input remains fixed, and no state reset or intermediate-answer input is used. Each map acts independently on all tokens. We also evaluate all combinations in which either map is omitted, giving nine two-loop conditions per fit.

This exploratory evaluation reuses the 512 training-excluded graphs from the target-switching experiment. All ten starting nodes give 5,120 examples. We require $u,f_G(u),\ldots,f_G^4(u)$ to be pairwise distinct, leaving the same 3,200 examples for every condition. We do not filter on prediction correctness. Requiring distinct labels through four hops makes this set smaller than those used for a single additional loop.

We define targets relative to $u$, not the model's first prediction. Each control that omits the second map is scored against the same cumulative target as the corresponding two-map sequence.

For both map fits, first-continuation predictions agree exactly with the earlier saved evaluations. Model parameters remain unchanged, and model, map, data, and code hashes are recorded. We bootstrap 5,000 paired samples of graph identities, preserving all starting nodes and both map fits within each sampled graph. Intervals quantify graph-sampling uncertainty for the two-fit mean on a fixed backbone.

\begin{center}\begin{minipage}{\linewidth}\centering
\captionsetup{hypcap=false}
\captionof{table}{Two-step composition accuracy (\%). Confidence intervals are 95\% graph-bootstrap intervals for the mean of the two map fits.}
\begin{tabular}{@{}llrrrr@{}}\toprule
Backbone seed & Sequence & Fit 1 & Fit 2 & Mean & 95\% CI\\\midrule
6 & One--one & 93.88 & 93.41 & 93.64 & [92.87, 94.36]\\
6 & One--two & 63.53 & 67.03 & 65.28 & [64.02, 66.50]\\
6 & Two--one & 73.22 & 75.78 & 74.50 & [73.42, 75.56]\\
6 & Two--two & 68.78 & 67.47 & 68.12 & [66.57, 69.71]\\
\bottomrule\end{tabular}\end{minipage}\end{center}

Composition accuracy depends on the sequence of maps. One--one reaches 93.64\% and two--two reaches 68.12\%. These results establish two-step reuse in seed 6 and motivate testing longer sequences.

\subsubsection{Eight-Step Controller Reuse}
\label{app:long-composition}

We reuse the same frozen maps over sequences of lengths 1--8 without further training. Starting from native $h_6$, we update the state as $h^{(j)}=F(J_{a_j}(h^{(j-1)}))$. We test repeated one-hop maps, repeated two-hop maps, and 32 distinct nonconstant binary strings of length eight. We sample the strings with a fixed seed before evaluation and use their prefixes for shorter sequences. The target at step $j$ is $f_G^{\sum_{i=1}^j a_i}(u)$, where $a_i\in\{1,2\}$. We never reset the state or supply an intermediate answer.

This exploratory evaluation uses the same 512 graphs and starting nodes as the two-step study. The primary set contains the same 3,200 instances with distinct $u,\ldots,f_G^4(u)$ at every sequence length. We do not filter on predictions, and both map fits exactly reproduce the saved two-step predictions.

A ten-node graph cannot have sixteen distinct successive nodes. Different cumulative targets can therefore coincide in the longer sequences. We report exact match between the predicted node and the cumulative target at each step. These tests measure repeated use on the same graph inputs; they do not test sixteen-hop extrapolation with distinct nodes at every step.

\begin{center}\begin{minipage}{\linewidth}\centering
\captionsetup{hypcap=false}
\captionof{table}{Exact match at the eighth additional controlled loop (percent, mean of two fits).}
\begin{tabular}{lrrr}\toprule
Seed & Repeated one & Repeated two & Mixed\\\midrule
6 & 9.66 & 10.62 & 10.11 \\
\bottomrule\end{tabular}\end{minipage}\end{center}

On the fixed 3,200-instance set, eighth-call exact match is approximately 10\% for each sequence family. Successful control over one or two loops therefore does not ensure that the resulting states remain controllable over longer sequences.

\begin{center}\begin{minipage}{\linewidth}\centering
\captionsetup{hypcap=false}
\captionof{table}{Accuracy at every controller-sequence length on the fixed 3,200-example population (percent, mean of two fits). Mixed averages the 32 fixed random strings.}

\begin{tabular}{@{}llrrrrrrrr@{}}\toprule
Seed & Sequence & 1 & 2 & 3 & 4 & 5 & 6 & 7 & 8\\\midrule
6 & one & 99.6 & 93.6 & 57.9 & 20.5 & 8.3 & 6.0 & 6.7 & 9.7 \\
6 & two & 98.0 & 68.1 & 23.4 & 10.8 & 11.6 & 11.1 & 10.4 & 10.6 \\
6 & mixed & 99.0 & 78.9 & 39.7 & 12.3 & 9.3 & 9.1 & 9.7 & 10.1 \\
\bottomrule\end{tabular}\end{minipage}\end{center}

\paragraph{Independent graph confirmation.}
We repeat the composition evaluation on 512 additional graphs sampled from the training-excluded locked pool with seed 2026093001. These graphs are disjoint from the 512 graphs used above. All ten starting nodes produce 5,120 examples; the same distinct-$u,\ldots,f_G^4(u)$ rule retains 3,175 examples, without filtering on predictions. We evaluate five existing D8L6 backbones and two existing map fits per backbone. No parameters are updated. The 32 mixed strings are fixed to those used in the earlier evaluation. Figure~\ref{fig:long-composition} shows this independent evaluation for seed 6, with 3,000 graph-bootstrap resamples for its intervals.

\begin{center}\begin{minipage}{\linewidth}\centering
\captionsetup{hypcap=false}
\captionof{table}{Independent composition evaluation (percent, mean of two fits). The final column gives the mixed-sequence target accuracy at the eighth additional call.}
\begin{tabular}{lrrrrr}\toprule
Seed & One--one & One--two & Two--one & Two--two & Mixed, call 8\\\midrule
6 & 93.64 & 65.04 & 75.13 & 68.03 & 10.09 \\
3 & 22.76 & 12.11 & 46.96 & 97.83 & 10.98 \\
4 & 83.15 & 41.81 & 56.57 & 28.79 & 9.05 \\
5 & 17.69 & 6.68 & 29.89 & 84.61 & 10.64 \\
7 & 19.80 & 9.15 & 26.61 & 91.72 & 11.57 \\
\bottomrule\end{tabular}\end{minipage}\end{center}

The independent results reproduce the dependence on map order and backbone. For seed 6, the four two-step accuracies are close to those in the original evaluation. At eight calls, mixed-sequence exact match is between 9.05\% and 11.57\% across the five backbones.

We also compare continuous control with omitting the current map after the identical controlled prefix, using the map only at the first call, and running the backbone without any map. We record the answer after $J$ but before $F$ and a baseline that simply copies $u$. The table below reports the mixed strings for seed 6 on the same 3,175 examples. Omitting only the second map reduces accuracy from 78.89\% to 32.93\%; reading directly after the second map gives 0.66\%. At longer lengths, continuous control no longer maintains an advantage. The copy baseline can score above zero when the requested walk returns to $u$; the recorded per-example results also support scoring only targets different from $u$.

\begin{center}\begin{minipage}{\linewidth}\centering
\captionsetup{hypcap=false}
\captionof{table}{Independent mixed-sequence controls for seed 6 (percent, mean of two fits). Columns count additional calls after $h_6$.}
\begin{tabular}{lrrr}\toprule
Condition & Call 2 & Call 4 & Call 8\\\midrule
Continuous $J$ & 78.89 & 12.62 & 10.09 \\
Omit current $J$ & 32.93 & 13.12 & 10.00 \\
$J$ at first call only & 32.93 & 7.52 & 14.99 \\
No $J$ & 2.93 & 10.47 & 16.18 \\
Read after $J$, before $F$ & 0.66 & 7.72 & 10.32 \\
Copy $u$ & 0.00 & 14.25 & 17.08 \\
\bottomrule\end{tabular}\end{minipage}\end{center}

\subsubsection{Selected Backbones and Aggregation}
\label{app:selected-mechanism}
We report mechanism experiments on three selected D8L6 backbones: seeds 6, 10, and 13. Seed 6 is the original case study. We selected seeds 10 and 13 from native readouts before evaluating their mechanism interventions. Both show one-hop and two-hop differences between consecutive model readouts. The mean across these backbones describes the selected cases, not a random sample of seeds.

The heatmaps use 512 held-out ten-node cycles and all ten starting nodes. They follow the visual conventions of Figure~\ref{fig:trajectory}. They show decoded predictions and do not directly measure internal operations.

Each backbone has two independently fitted rank-48 maps per target, trained as described in Appendix~\ref{app:target-setup}. We average the fits within each backbone, then average the three backbone accuracies with equal weights. We do not pool examples across backbones. Figure~\ref{fig:parallel-mechanism}b,c shows each backbone mean. Figure~\ref{fig:target-pattern-patching} shows their range, which is not a confidence interval.

For seeds 10 and 13, we select heads using 32 independent graph pairs. The selected second-layer heads are H0 and H3, respectively; seed 6 retains H2. We evaluate all four heads on confirmation data. The new confirmation set contains 512 graphs in each run. A separate set of 512 graphs is used for second-layer target-switching patches.

All graph identities come from the reserve excluded from backbone and map training. Discovery and confirmation sets are disjoint. The new backbones share these sets, while seed 6 retains its original independent sets. Input alignment, interventions, and inclusion rules remain unchanged. Patches between steered and unsteered runs replace the patterns of all four second-layer heads at the answer position. Target-switching patches cover all heads and positions in layer 2.

The cross-graph eligible counts for fits 1/2 are 3,966/3,963 (seed 6), 4,104/4,104 (seed 10), and 3,822/3,692 (seed 13). Patching between steered and native runs uses 4,044 instances for seed 6 and 4,111 for each new backbone. Target switching uses 4,116 and 4,149, respectively. The latter two experiments exclude label collisions without conditioning on correct predictions.

We also replace queries and restore outputs using the same protocol and selected head for each backbone. The comparison head is the next head modulo four. Averaging within and then across backbones gives 74.0\% recovery for the selected head, 0.5\% for the comparison head, and 100\% for all four heads. Recovery is measured among eligible answers that are correct before query replacement and incorrect afterward.

\section{Ouro Letter-Walk}
\label{app:parallel}
This appendix describes the language-based graph task and attention interventions used in Section~\ref{sec:hypothesis}.

\subsection{Task Definition and Input Format}
Letter-walk expresses a graph walk in text and has a known answer at each requested depth.

Each example defines a cycle over ten people using ten sentences of the form ``Alice passes the letter to Bob.'' The input also names the starting person and the number of steps to follow. Names are Alice, Bob, Carol, David, Emma, Frank, Grace, Henry, Iris, and Jack. Rules appear in shuffled order, so their textual order does not give the path. Ouro's tokenizer and chat template encode the instructions, rule sentences, and question as a user message. The target is an assistant message containing only the final person's name.

For example, take the cycle Alice $\to$ Bob $\to$ Carol $\to$ David $\to$ Emma $\to$ Frank $\to$ Grace $\to$ Henry $\to$ Iris $\to$ Jack $\to$ Alice. The prompt includes all ten rules. Starting with Alice, the two-step answer is \texttt{Carol} and the eight-step answer is \texttt{Iris}. For illustration, the question can be phrased as:
\begin{quote}
The letter starts with Alice. After exactly 02 steps, who has it? Answer directly with only the person's name.
\end{quote}
This example paraphrases the question wording; \texttt{02} preserves the template's two-digit encoding of the requested step count. The rules and question are ordinary text, not one special token per person or per edge. Different prompt templates express the same task.

\subsection{Model and Steering Configuration}
\label{app:ouro-config}

\paragraph{Supervision and steering.}
We fine-tune the backbone on 1--4-step requests with four recurrent loops. Answer-token cross-entropy is applied only at loop 4. Prompt tokens are masked from this loss. The target contains the assistant's answer and the chat-template ending, with no intermediate reasoning trace.

We then freeze the backbone and fit a dense affine map on 1--8-step requests. The same map acts on every token before loops 2--4, and answer supervision remains at loop 4. Both training stages combine answer cross-entropy and general-text cross-entropy with weights 0.8 and 0.2. An eight-step request is therefore beyond the backbone-training depths but within the map-training range. All mechanism evaluations request eight steps. Patching changes internal activations while preserving the requested task and answer rule.

Backbone fine-tuning uses Adafactor with factored FP32 states and no first moment, learning rate $10^{-5}$, and 50 warmup updates. Each update contains eight task examples. Map fitting uses AdamW, learning rate $10^{-4}$, 50 warmup updates, and sixteen task examples per update, with two examples for each requested depth 1--8. Both stages use FP32 weights with BF16 autocast and gradients through all four loops. The early-exit gate is frozen and unused. The dense map starts as the identity and has 4,196,352 parameters.

\paragraph{Checkpoint and intervention sites.}
All Ouro mechanism tests use the letter-walk Ouro-2.6B checkpoint saved after 200 backbone updates and its dense affine map saved after 500 map updates. The frozen map is $J(h)=hW+b$ with unrestricted $W$, applied to all token states before loops 2--4. Head outputs are patched before their output projection during prompt processing; generation follows the normal steering schedule of the patched run.

\subsection{Evaluation and Attention Interventions}
\label{app:ouro-evaluation}
We keep the trained backbone and map fixed during head selection and evaluation. Each intervention experiment uses a separate evaluation set.

\paragraph{Head selection.}
The selected heads are L41.H2/H10/H12/H15, L43.H1/H6/H11/H13/H15, and L47.H0/H2/H3/H4/H7/H13/H15, all at loop 4. Layer and head indices start at zero. We select heads using eight discovery pairs and a separate 64-pair confirmation set for pattern patching. We then fix these sixteen heads. The pattern-versus-output, restoration, and steering-pattern evaluations each use a separate set of 256 pairs.

The pattern-versus-output and restoration experiments use sixteen comparison heads, matched by layer to the selected heads.

\paragraph{Restoring outputs after query replacement.}
The restoration experiment uses 256 new query pairs generated with seed 2026092642. The graph pairs use the same seven-edge shared-path construction as the pattern-versus-output experiment. However, replacement queries come from a run on the same graph with a different starting person. This evaluation set excludes all earlier evaluation graphs, including the new pattern-versus-output set.

At loop 4, we replace all queries in layers 41, 43, and 47. The patched run keeps its own keys and values. We then restore the clean outputs of the sixteen selected heads, sixteen comparison heads matched by layer, or all 48 affected heads. We measure recovery only for answers that were correct before query replacement and incorrect afterward. Restoration outcomes do not determine which examples are included.

\paragraph{Pairs that separate routing from retrieved content.}
The pattern-versus-output experiment uses 256 new pairs generated with seed 2026092641. Their 512 distinct graph identities exclude all 1,184 identities from earlier head selection and mechanism evaluations, including the new steering-pattern set. Each graph is a ten-person cycle. From the starting person in run 1, the first seven people on the path are the same in both graphs, but the eighth differs. Run 2 starts at a different person. Names occupy aligned token positions.

The pair construction ensures three distinct answers: the original answer in run 2, the answer in run 1, and the answer in the second graph from the starting person used in run 1. We evaluate the third answer with a reference run that combines that graph and starting person. Pair inclusion does not depend on model accuracy. This construction defines the predicted answers under pattern and output patching without assuming one graph step per loop.

\paragraph{Independent steering-pattern evaluation.}
We sample 256 new graph pairs with seed 2026092632. Their 512 distinct graph identities exclude all 672 identities from the preceding mechanism evaluations and head selection. Each graph is a ten-person cycle. The paired prompts have aligned token positions, and the existing pair-construction rules are unchanged. No pair is filtered by model correctness. The backbone, map, sixteen heads, and loop-4 intervention sites remain fixed; no training or head reselection is performed.

For the same input, we patch steered patterns into an unsteered run and unsteered patterns into a steered run. The controls patch only the steered values, patterns from an unrelated graph, loop-2 patterns into loop 4, or steered patterns at sixteen disjoint comparison heads matched by layer. These comparison heads are L41.H0/H1/H3/H4, L43.H0/H2/H3/H4/H5, and L47.H1/H5/H6/H8/H9/H10/H11.

All patches affect prompt processing only. During generation, each run keeps its original steering schedule. We use greedy generation with a maximum of sixteen tokens and score both the first answer token and the complete name.

\paragraph{Numerical checks and statistical reporting.}
Same-run output replacement and restoration of every affected output both yield zero maximum logit error. In the steering-pattern evaluation, same-run output replacement also gives zero maximum logit error. Explicit same-run pattern reconstruction differs from BF16 attention by at most 0.125 in logits and yields no accuracy gain (0/256).

The Ouro figures report Wilson confidence intervals. When full names are generated, we compare their correctness with first-token correctness and count truncated generations as incorrect. No verified graph-exclusion list is available for the backbone. We therefore cannot establish that these graph identities were absent from backbone training.

\subsection{Additional Results}
\label{app:ouro-results}

\paragraph{Head-output restoration.}
The table reports both accuracy over all query pairs and recovery among clean answers broken by query replacement.
\begin{center}\begin{minipage}{\linewidth}\centering\centering
\captionsetup{hypcap=false}
\captionof{table}{Restoring Ouro head outputs on 256 new query pairs. Recovery conditions on clean answers broken by query replacement.}
\begin{tabular}{@{}lrr@{}}\toprule
Condition & Correct / 256 & Recovered / broken\\\midrule
Clean & 251 & --- \\
Replaced queries & 2 & 0/249 \\
Selected 16 heads & 251 & 249/249 \\
Matched 16 neighbors & 2 & 0/249 \\
All 48 affected heads & 251 & 249/249 \\
\bottomrule\end{tabular}\end{minipage}\end{center}

The selected outputs recover 100.0\% of eligible broken answers (Wilson 95\% interval 98.5--100.0\%). Complete-name correctness agrees with first-token correctness for the clean, corrupted, selected-head, comparison-head, and all-head conditions. No generation is truncated, and restoring all affected outputs exactly reconstructs the clean logits.

\paragraph{Pattern-versus-output patching.}
The table reports the answers from each unpatched run and the answer predicted by pattern patching.
\begin{center}\begin{minipage}{\linewidth}\centering\centering
\captionsetup{hypcap=false}
\captionof{table}{Ouro pattern-versus-output patching on 256 new pairs. Counts use first-answer-token predictions. Patches use activations from run 1 in run 2. ``Rerouted'' denotes the answer in the second graph under the patched pattern.}
\begin{tabular}{@{}lrrrr@{}}\toprule
Condition & Run 2 & Run 1 & Rerouted & Other\\\midrule
Run 2 & 252 & 0 & 0 & 4 \\
Run 1 & 0 & 252 & 2 & 2 \\
Graph from run 2, start from run 1 & 0 & 4 & 251 & 1 \\
16-head pattern & 37 & 0 & 182 & 37 \\
16-head output & 5 & 240 & 7 & 4 \\
16-head values & 221 & 0 & 24 & 11 \\
Matched-neighbor pattern & 251 & 0 & 0 & 5 \\
Matched-neighbor output & 252 & 0 & 0 & 4 \\
Wrong-loop pattern & 137 & 3 & 0 & 116 \\
\bottomrule\end{tabular}\end{minipage}\end{center}

Pattern patching produces the answer predicted by the patched pattern in 182/256 cases (71.1\%; Wilson 95\% interval 65.3--76.3\%). It produces the answer from run 1 in 0/256 cases. Output patching produces these two answers in 7/256 and 240/256 cases, respectively. These results distinguish changing where attention reads from replacing the retrieved output.

Full-name and first-token classifications agree on all 256 pairs in every generated condition. No generation is truncated, so the distinction holds under both scoring rules.

\paragraph{Steering-pattern patching.}
The earlier 64-pair evaluation gave 61/64 correct answers under the sixteen-head pattern patch, compared with 62/64 under full steering and 0/64 without steering. Table~\ref{tab:ouro-steering-256} reports the independent 256-pair evaluation used in Figure~\ref{fig:parallel-mechanism}c.

On this new set, the fixed pattern patch reaches 248/256 correct answers (96.9\%; Wilson 95\% interval 94.0--98.4\%). Full steering reaches 251/256. The reverse pattern patch and all four comparison interventions score zero. Full-name and first-token correctness agree for every example in all eight conditions, and no generation is truncated.

\begin{center}
\begin{minipage}{\linewidth}
\centering
\captionsetup{hypcap=false}
\captionof{table}{Independent Ouro steering-pattern evaluation with the fixed sixteen heads. First-token and complete-name scores agree in all conditions.}
\label{tab:ouro-steering-256}
\begin{tabular}{@{}lrr@{}}\toprule
Condition & Correct / 256 & Accuracy (\%)\\\midrule
Unsteered & 0 & 0.0 \\
Full steering & 251 & 98.0 \\
Steered patterns $\to$ unsteered & 248 & 96.9 \\
Unsteered patterns $\to$ steered & 0 & 0.0 \\
Steered values only $\to$ unsteered & 0 & 0.0 \\
Unrelated-graph patterns $\to$ unsteered & 0 & 0.0 \\
Wrong-loop patterns $\to$ unsteered & 0 & 0.0 \\
Comparison-head patterns $\to$ unsteered & 0 & 0.0 \\
\bottomrule\end{tabular}
\end{minipage}
\end{center}

Full steering succeeds on four examples where the pattern patch fails. The pattern patch succeeds on one example where full steering fails. The paired accuracy gap is 1.17 percentage points (95\% bootstrap interval $-0.39$ to $2.73$, using 10,000 resamples of graph pairs). Thus, the fixed pattern patch recovers the steering effect on new evaluation graphs. These results do not establish equivalence to full steering or replication across independently trained Ouro backbones.

\section{Matched Supervision in Graph Walk}
\label{app:matched-supervision}
This experiment tests the effect of adding intermediate losses while keeping the query and backbone training stream fixed. It uses ten-node D8L8 models trained separately from the D8L6 control models and the eight-node continuation models.

\subsection{Paired Backbone Training}

Each model repeats a two-layer Transformer block for eight loops. The hidden dimension is 256, with four attention heads, MLP dimension 1,024, and no dropout. We train five pairs with seeds 0--4. Within each pair, both models start from identical parameters and receive the same sequence of training tokens. Every input requests the eight-hop target.

Let $\ell_t$ be cross-entropy on $f_G^t(s)$ at loop $t$. The final-only objective is $\ell_8$. The stepwise objective is
\begin{equation}
\mathcal{L}_{\mathrm{stepwise}}=\ell_8+\frac{1}{7}\sum_{t=1}^{7}\ell_t.
\end{equation}
Thus, the final-loop coefficient remains one. Adding the intermediate losses changes the total loss and gradient scale; these quantities are not normalized to match final-only training.

Both regimes use 20,000 AdamW updates, batch size 512, learning rate $3\times10^{-4}$, weight decay 0.3, 500 warmup updates, and cosine learning-rate decay. Gradients pass through all eight loops. We use the final checkpoint of every run. Saved hashes verify that each pair shares its initialization and complete training-token stream.

\subsection{Maps and Evaluation}

We freeze each backbone and collect $h_8$ on fixed depth-eight queries. Three separately fitted maps target $u=f_G^8(s)$, $f_G(u)$, or $f_G^2(u)$ after one additional call to $F$. Each map has the form $J(h)=h(D+AB)+b$, with a learned diagonal $D$, rank 48, and 25,088 parameters. It acts on all tokens and starts as the identity.

Each target has two map fits. Each fit uses 8,000 AdamW updates, batch size 128, learning rate $10^{-4}$, zero weight decay, and gradient clipping at 1.0. We evaluate on the selection set every 400 updates and retain the earliest checkpoint with the highest selection accuracy. The map-training, selection, and test pools contain 2,048, 512, and 512 graphs, respectively. The pools are disjoint and shared across all ten backbones. The test graphs are excluded from backbone training.

The test pool contains 5,120 graph--start instances. We use the same 4,137 instances with distinct $u$, $f_G(u)$, and $f_G^2(u)$ in every condition. Inclusion depends on graph structure, not predictions. All accuracies below are read after the additional frozen loop. We average the two fits within a backbone before computing the mean and sample standard deviation across the five backbones.

\subsection{Results across Paired Seeds}

Without a map, all five final-only backbones retain $u$, and all five stepwise backbones predict $f_G(u)$ on the scoring population. Every fitted stay map reaches 100\% accuracy in both regimes. Every stepwise one-hop map also reaches 100\%. Table~\ref{tab:matched-supervision} summarizes all conditions, and Table~\ref{tab:matched-fits} reports the remaining individual fits.

\begin{figure}[!htb]
\centering
\includegraphics[width=\linewidth]{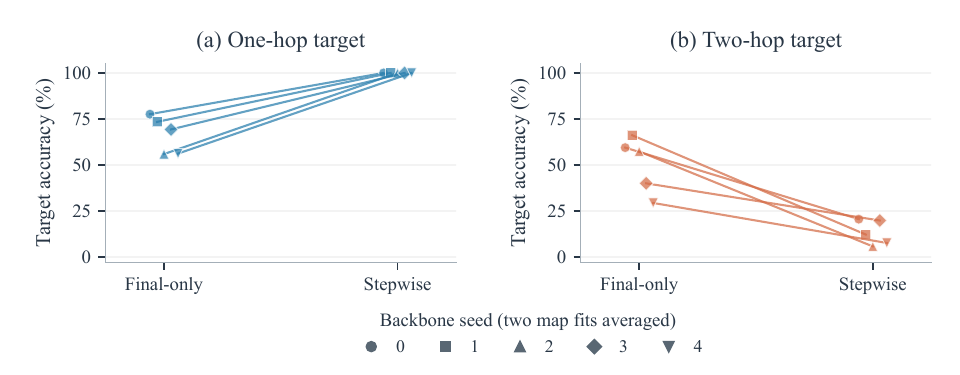}
\caption{Paired backbone comparisons for (a) one-hop and (b) two-hop targets. Lines connect the two supervision regimes for the same backbone seed. Points average two map fits; horizontal offsets separate overlapping points.}
\label{fig:matched-supervision-pairs}
\end{figure}

Figure~\ref{fig:matched-supervision-pairs} shows the one-hop and two-hop comparisons within each backbone pair.

\begin{table}[!htb]
\centering
\caption{Matched ten-node D8L8 supervision comparison. Entries are mean target accuracies $\pm$ sample standard deviation across five backbone seeds (\%). We first average the two map fits within each backbone. Under $J$, each column uses its own target-specific map.}
\label{tab:matched-supervision}
\begin{tabular}{@{}llrrr@{}}\toprule
Supervision & Intervention & Stay & One hop & Two hops\\\midrule
Final-only & None & $100.0\pm0.0$ & $0.0\pm0.0$ & $0.0\pm0.0$\\
Stepwise & None & $0.0\pm0.0$ & $100.0\pm0.0$ & $0.0\pm0.0$\\
Final-only & $J$ & $100.0\pm0.0$ & $66.5\pm10.0$ & $50.5\pm15.2$\\
Stepwise & $J$ & $100.0\pm0.0$ & $100.0\pm0.0$ & $13.2\pm6.8$\\
\bottomrule\end{tabular}
\end{table}

\begin{table}[!htb]
\centering
\caption{Matched D8L8 map accuracy (\%) on 4,137 common test instances. Each pair of columns reports the two independently fitted maps for that target.}
\label{tab:matched-fits}
\begin{tabular}{@{}llrrrr@{}}\toprule
 & & \multicolumn{2}{c}{One hop} & \multicolumn{2}{c}{Two hops}\\
Seed & Supervision & Fit 1 & Fit 2 & Fit 1 & Fit 2\\\midrule
0 & Final-only & 78.75 & 76.46 & 62.05 & 56.80\\
0 & Stepwise & 100.00 & 100.00 & 25.57 & 15.54\\
1 & Final-only & 74.09 & 72.71 & 66.76 & 65.55\\
1 & Stepwise & 100.00 & 100.00 & 11.58 & 12.79\\
2 & Final-only & 56.42 & 55.43 & 57.77 & 56.90\\
2 & Stepwise & 100.00 & 100.00 & 6.48 & 5.00\\
3 & Final-only & 68.41 & 70.22 & 39.21 & 40.83\\
3 & Stepwise & 100.00 & 100.00 & 19.17 & 20.45\\
4 & Final-only & 56.78 & 55.60 & 30.67 & 28.04\\
4 & Stepwise & 100.00 & 100.00 & 7.57 & 7.69\\
\bottomrule\end{tabular}
\end{table}

\FloatBarrier
One-hop accuracy is $66.5\pm10.0\%$ for final-only models and $100.0\pm0.0\%$ for stepwise models. Two-hop accuracy reverses this ordering: $50.5\pm15.2\%$ for final-only models and $13.2\pm6.8\%$ for stepwise models. Here $\pm$ denotes the sample standard deviation across the five backbone means. Final-only two-hop accuracy is higher in every pair. These results describe the transitions reached by the tested map family and fitting procedure. They do not establish the limits of other state interventions.

\FloatBarrier
\section{Ouro Antonym Cancellation}
\label{app:supervision}
This appendix defines the task and the two supervision strategies compared in Section~\ref{sec:supervision-boundary}.

\subsection{Task Definition and Input Format}

The input contains a word sequence, a deletion rule, and a requested deletion count. At each step, the rule removes the leftmost adjacent antonym pair and joins the remaining words in their original order. The vocabulary consists of twelve pairs: hot/cold, near/far, open/closed, heavy/light, young/old, rich/poor, happy/sad, clean/dirty, full/empty, wet/dry, fast/slow, and alive/dead. Actual examples contain all 24 words, with pair orientations and nesting varied. Ouro encodes the full text using its tokenizer and chat template. The answer contains the two words removed at the requested step, in their left-to-right order.

A short illustrative sequence is \texttt{hot near far cold open closed}. The first deletion removes \texttt{near far}, leaving \texttt{hot cold open closed}. The second removes \texttt{hot cold}, and the third removes \texttt{open closed}. Thus, the answer to deletion 2 is \texttt{hot cold}, not the remaining sequence. This shortened example illustrates the rule; training and evaluation use 24-word inputs. In the final-only format, the question is ``Which two words are deleted on deletion number 2? Answer with only those two words in their original left-to-right order.'' The stepwise format instead states the requested number of deletions and asks for the pair removed in the current deletion update.

\subsection{Model and Steering Configuration}

\paragraph{Answer-token supervision.}
The user prompt is masked from the task loss. Cross-entropy supervises the assistant's two-word answer and chat-template ending, without a generated deletion trace.

\paragraph{Backbone supervision.}
We fine-tune the two Ouro backbones separately for 500 updates, with four loops in every run. Final-only training samples requests for deletions 1--4 and supervises the requested pair at loop 4. Stepwise training always requests four deletions. It supervises the pair removed at deletion $t$ after loop $t$, for $t=1,\ldots,4$. Thus, one model learns to answer a requested depth at the final loop, while the other learns a fixed sequence of intermediate answers.

\paragraph{A shared map for each frozen backbone.}
Each backbone receives its own map $J(h)=h(I+AB)+b$, with a rank-128 residual update and 526,336 trainable parameters. It is shared across all four loops and acts on every token after each loop's RMSNorm, including after loop 4 before the frozen language-model head. Zero initialization of $B$ and $b$ makes the initial map the identity.

Both fits use 500 AdamW updates at learning rate $10^{-4}$, with 50 warmup updates. An update contains sixteen examples, two for each requested deletion depth 1--8. The loss weights answer cross-entropy by 0.8 and general-text cross-entropy by 0.2. During both map fitting and evaluation, the prompt specifies the requested deletion count. For both backbones, map training supervises the requested pair at the fixed loop-4 exit. Adaptive halting is disabled so all requests use four loops.

\subsection{Evaluation Protocol}

\paragraph{Native readout heatmaps.}
Figure~\ref{fig:ouro-native-readouts}a,b uses the unsteered backbones at update 500. In the stepwise panel, every prompt requests four deletions. For each of 64 sequences, we compare the parsed pair at each loop with all four true deletion pairs to construct the depth-by-loop matrix.

In the final-only panel, each prompt requests the depth shown on the vertical axis. Each cell contains 64 examples from the saved loop-1/2 and loop-3/4 evaluations. Both panels score the parsed deletion pair, without requiring exact full-text generation or EOS emission.

\paragraph{Paired evaluation at deeper requested deletions.}
The main evaluation uses prompts that specify a deletion count and ask for the pair removed at that step (template 0 in the implementation). At each depth, the same 64 held-out sequence structures and target pairs are evaluated in all four conditions: each backbone with and without its map. Success requires matching the requested deletion pair. Figure~\ref{fig:ouro-native-readouts}c reports the evaluation at map update 500.

At deletion depths 5--8, final-only plus $J$ answers 251/256 requests correctly and stepwise plus $J$ answers 20/256. At depth 8, the counts are 63/64 and 2/64. Both native models score zero on depths 5--8. In the native final-only readout panel, loop 3 answers 255/256 requests correctly, including 63/64 at deletion 4.

We evaluate one backbone and one map per training strategy. Depths 5--8 are beyond backbone training but are included in map training. The low stepwise accuracy shows that the tested map and fitting procedure fail on these requests. It does not rule out other state interventions that could produce the correct answers.

\FloatBarrier

\section{Training and Intervention Summary}
\label{app:training-summary}
Table~\ref{tab:training-summary} summarizes the training budgets and intervention schedules for the main-text experiments. The corresponding appendices specify losses, sampling rules, optimizer settings, and checkpoint selection. Repeated map fits optimize separate maps on a frozen backbone; they are not independent backbone runs.

\begin{table}[!htbp]
\centering
\caption{Training budgets and intervention sites for the main-text experiments. Backbone updates identify the selected checkpoint or the training budget as specified; map updates give the fitting budget.}
\label{tab:training-summary}
\setlength{\tabcolsep}{3pt}
\begin{tabular}{@{}>{\raggedright\arraybackslash}p{.22\linewidth}>{\raggedright\arraybackslash}p{.24\linewidth}>{\raggedright\arraybackslash}p{.18\linewidth}>{\raggedright\arraybackslash}p{.28\linewidth}@{}}\toprule
Experiment & Backbone & Map & Placement\\\midrule
D8L6 control & 20,000 updates; seed 6 selected at 16,000 & Rank 48; 8,000 updates; two fits per target & All tokens before the next $F$; maps reused in compositions.\\
Matched graph supervision & 20,000 updates; five paired final checkpoints & Rank 48; 8,000 updates; two fits per target & At $h_8$, followed by one additional $F$.\\
Ouro letter-walk & Checkpoint at update 200 & Dense affine; checkpoint at update 500 & Before loops 2--4; all tokens.\\
Ouro cancellation & 500 updates per training strategy & Residual rank 128; 500 updates & After each loop's RMSNorm, including loop 4 before the output head.\\
\bottomrule\end{tabular}
\end{table}

\FloatBarrier
\clearpage
\section{Additional Evidence for State Steering}
\label{app:additional-control}
We provide two further demonstrations of state steering: numerical and calendar tasks with a task-adapted Qwen3-8B model, and relation composition in a synthetic knowledge graph. In both settings, a shared affine map modifies hidden states between calls to a frozen backbone and substantially improves the model's ability to reach the requested target.

\subsection{Control of Continued Computation in Qwen3-8B}
\label{app:qwen-control}

\paragraph{Tasks and model.}
The five task families are integer successor, weekday advancement, repeated doubling, Fibonacci-pair updates, and Collatz iteration. Each prompt specifies an initial state and an iteration count $k$, and asks for the final answer directly. For Fibonacci, an update maps $(a,b)$ to $(b,a+b)$ and the answer is the second component after $k$ updates. For example, a doubling prompt starting at 8 with $k=3$ has answer 64.

We form a recurrent model by sharing Qwen3-8B's full stack of 36 decoder layers across loops. Token embeddings enter once, the native token-position RoPE is retained across loops, and the final RMSNorm and language-model head are applied after the last loop. The backbone is first adapted with four loops and requests $k=1,\ldots,4$, supervising the final answer. We then freeze the backbone and train one token-wise dense affine map $J(h)=h+Wh+b$, shared across all five tasks and all seven boundaries of an eight-loop execution. The map has 16,781,312 parameters, starts from the identity, and is trained on requests $k=1,\ldots,8$ using final-answer cross-entropy. The evaluated backbone and map are the checkpoints at updates 9,500 and 12,000, respectively.

\paragraph{Paired continuation comparison.}
The evaluation contains 32 base cases per task and all eight requested counts, giving 1,280 prompts. Every branch preserves the controlled first four loops. From this common boundary $h_4$, we compare four suffixes: four further $J$--$F$ pairs, four $F$ calls without $J$, four applications of $J$ without $F$, and immediate readout. Thus, the $J+F$ and $F$-only branches both use eight backbone calls in total. Answers are generated greedily, without a KV cache, with a 12-token limit. Accuracy requires the complete generated answer to match the target after trimming surrounding whitespace and ignoring case. Each branch generates its own continuation from the same prompt.

\begin{table}[!htbp]
\centering
\caption{Qwen continuation accuracy (\%). All branches share the controlled computation through $h_4$; column headings specify what happens afterward. Each task row contains 128 long requests ($k=5$--8). The pooled short and long rows each contain 640 prompts.}
\label{tab:qwen-additional-control}
\begin{tabular}{@{}lrrrr@{}}\toprule
Task or request range & $J+F$ & $F$ only & $J$ only & Stop\\\midrule
Successor & 100.00 & 75.00 & 71.09 & 70.31\\
Weekday & 100.00 & 61.72 & 78.12 & 52.34\\
Doubling & 100.00 & 3.91 & 4.69 & 3.12\\
Fibonacci & 100.00 & 0.00 & 0.00 & 0.00\\
Collatz & 48.44 & 2.34 & 2.34 & 2.34\\\midrule
All long requests ($k=5$--8) & 89.69 & 28.59 & 31.25 & 25.62\\
All short requests ($k=1$--4) & 99.38 & 100.00 & 100.00 & 100.00\\
\bottomrule\end{tabular}
\end{table}

\paragraph{Results.}
Continuing with $J+F$ answers 574/640 long requests correctly, compared with 183/640 for continued $F$ alone, 200/640 for continued $J$ alone, and 164/640 for stopping at $h_4$ (Table~\ref{tab:qwen-additional-control}). The full controlled execution reaches 100\% on successor, weekday, doubling, and Fibonacci, and 48.44\% on Collatz. Removing either the state intervention or the backbone from the suffix sharply reduces long-request accuracy. These results show that a shared state intervention can support effective continued computation across several task families in the same frozen language model.

\FloatBarrier
\clearpage
\subsection{Controlled Relation Composition in a Synthetic Knowledge Graph}
\label{app:kg-control}

\paragraph{Task and frozen backbone.}
The knowledge graph contains 64 entities and 16 relations. Each relation $r$ is a fixed permutation $f_r$ of the entity set. Given a starting entity $e_0$ and a relation sequence $(r_1,\ldots,r_n)$, the target is
\begin{equation}
e_n=f_{r_n}\circ\cdots\circ f_{r_1}(e_0).
\end{equation}
Inputs consist of a beginning-of-sequence token, the initial entity identifier, and the relation identifiers. The same relation permutations are used throughout training and evaluation. The backbone has a shared two-layer Transformer block, width 256, eight attention heads, MLP width 1,024, causal attention, and no positional encoding. Embeddings enter only at the first loop. Backbone training uses relation lengths 1--3 with cross-entropy on the aligned intermediate entities; we use the selected checkpoint at update 16,000.

\paragraph{Affine state control.}
After freezing the backbone, we fit an identity-initialized, token-wise affine map with 65,792 parameters. A length-$n$ query uses one initial $F$ call followed by $n-1$ repetitions of $J$ then $F$. The same map is reused at every boundary. The comparison without $J$ uses the same input, frozen backbone, readout, and $n$ calls to $F$.

The reported map first follows a length curriculum through every integer length from 4 to 16, with up to 3,000 updates per stage, batch size 128, and learning rate $10^{-4}$. Training supervises the correct intermediate entity after each $F$ call from call 2 onward. Starting from this length-16 checkpoint, a further 10,000 updates mix 5,000 batches with lengths sampled uniformly from 4--16 and 5,000 batches at length 16. This continuation uses AdamW with zero weight decay and a learning rate that rises to $3\times10^{-5}$ over 500 updates, stays constant for 8,500 updates, and decays to zero over the last 1,000. We evaluate the final checkpoint.

\begin{figure}[!htbp]
\centering
\includegraphics[width=.88\linewidth]{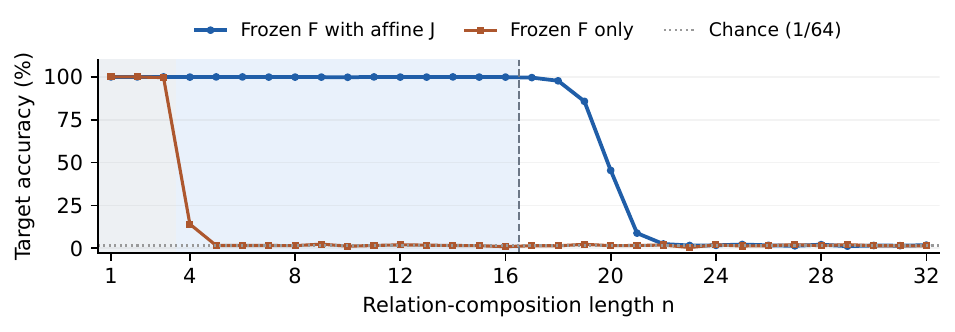}
\caption{State steering in synthetic KG relation composition. Each point uses 1,024 queries, paired between the two conditions, with exactly $n$ backbone calls for a length-$n$ query. Gray shading marks backbone training lengths 1--3; blue shading marks map training lengths 4--16. The dashed vertical line marks the end of map training coverage.}
\label{fig:kg-additional-control}
\end{figure}

\paragraph{Results.}
We evaluate every length from 1 to 32 on 1,024 sampled queries per length. Across lengths 4--16, the affine map raises target accuracy from 2.53\% to 99.92\% (Table~\ref{tab:kg-additional-control}). Accuracy with $J$ remains 99.61\% at length 17, 97.75\% at length 18, and 85.74\% at length 19, before declining with longer compositions (Figure~\ref{fig:kg-additional-control}). The average is 42.87\% over lengths 17--24 and 1.65\% over lengths 25--32. The affine intervention therefore enables reliable relation composition well beyond the backbone's original training lengths, with a clear decline as execution moves farther beyond the map's training range.

\begin{table}[!htbp]
\centering
\caption{KG target accuracy aggregated over equally sized per-length evaluation sets. Every row compares the same queries and the same number of frozen-backbone calls.}
\label{tab:kg-additional-control}
\begin{tabular}{@{}lrrr@{}}\toprule
Relation length & Queries & $F$ only (\%) & With affine $J$ (\%)\\\midrule
1--3 & 3,072 & 99.87 & 100.00\\
4--16 & 13,312 & 2.53 & 99.92\\
17--24 & 8,192 & 1.54 & 42.87\\
25--32 & 8,192 & 1.56 & 1.65\\
\bottomrule\end{tabular}
\end{table}

\FloatBarrier
Together, the Qwen and KG experiments provide additional evidence that shared affine interventions at loop boundaries can substantially improve controlled execution with frozen backbones, extending the state-steering results to a task-adapted language model and a learned symbolic relation executor.

\end{document}